\documentclass[11pt]{article}

\usepackage[preprint]{acl}

\usepackage{times}
\usepackage{latexsym}
\usepackage{multirow}
\usepackage[T1]{fontenc}

\usepackage[utf8]{inputenc}

\usepackage{microtype}

\usepackage{inconsolata}

\usepackage{graphicx}
\usepackage{bbm}
\usepackage{amsmath}
\usepackage{amsfonts}
\usepackage{amssymb}
\usepackage{amsthm}
\usepackage{dsfont}
\usepackage{caption}
\usepackage{subcaption}
\usepackage{soul}
\usepackage{hyperref}
\usepackage{adjustbox}
\usepackage{booktabs}
\usepackage{makecell}
\usepackage{nicefrac}
\usepackage{comment}
\usepackage{rotating}
\usepackage{scalerel}
\usepackage{lipsum}
\usepackage{svg}
\usepackage{mathtools}
\usepackage{mleftright}
\usepackage{booktabs}
\usepackage{comment}
\usepackage{xspace}
\usepackage{linguex}
\usepackage[dvipsnames]{xcolor} 

\usepackage[textsize=tiny]{todonotes}

\definecolor{darkcerulean}{rgb}{0.03, 0.27, 0.49}

\definecolor{bronze}{rgb}{0.8, 0.5, 0.2}
\definecolor{ga}{rgb}{0.451, 0.667, 0.698}
\definecolor{gd}{rgb}{0.780, 0.565, 0.063}

\newcommand{\goaldirected}[1]{\textcolor{gd}{#1}}
\newcommand{\goalagnostic}[1]{\textcolor{ga}{#1}}

\definecolor{Blue}{RGB}{33,92,175}   
\definecolor{Green}{RGB}{98,115,19}      
\definecolor{PurpleDark}{RGB}{140,10,89} 
\definecolor{Purple}{RGB}{163,7,116} 
\definecolor{Gray}{RGB}{111,111,111} 
\definecolor{Red}{RGB}{183,53,45}   
\definecolor{Petrol}{RGB}{0,120,148}
\definecolor{Bronze}{RGB}{142,103,19} 
\definecolor{Orange}{RGB}{230, 100, 50}

\definecolor{YlOrBr3}{HTML}{FEC44F}
\definecolor{YlOrBr4}{HTML}{FEB23C}
\definecolor{YlOrBr5}{HTML}{FE9929}
\definecolor{YlOrBr6}{HTML}{EC7014}
\definecolor{YlOrBr7}{HTML}{CC4C02}
\definecolor{YlOrBr8}{HTML}{993404}
\definecolor{YlOrBr9}{HTML}{662506}

\colorlet{MacroColor}{Green}
\colorlet{MACROCOLOR}{MacroColor}

\usepackage[capitalize,noabbrev]{cleveref}
\crefname{section}{\S}{\S\S}
\crefname{table}{Tab.}{Tab.}
\crefname{figure}{Fig.}{Figs.}
\crefname{algorithm}{Alg.}{}
\crefname{equation}{Eq.}{Eq.}
\crefname{appendix}{App.}{App.}
\crefname{theorem}{Theorem}{}
\crefname{restatableTheorem}{Theorem}{}
\crefname{prop}{Proposition}{}
\crefname{definition}{Def.}{}
\crefname{cor}{Corollary}{}
\crefname{observation}{Observation}{}
\crefname{assumption}{Assumption}{}
\crefname{hyp}{Hyp.}{Hypotheses}
\crefformat{section}{\S#2#1#3}
\crefname{namedtheorem}{Hyp.}{Hypotheses}

\newcommand{\defn}[1]{\textbf{#1}}

\newcommand{\word}[1]{\emph{#1}}

\newcommand{\defeq}{\;\overset{\text{\tiny def}}{=}\;}
\newcommand{\defpropto}{\;\mathrel{\stackrel{\textnormal{\tiny def}}{\propto}}\;}

\newcommand{\alphabet}{{\Sigma}}
\newcommand{\kleene}[1]{{#1^{*}}}

\newcommand{\strings}{{\kleene{\alphabet}}}

\newcommand{\str}{{\boldsymbol{w}}}
\newcommand{\sym}{{w}}

\newcommand{\contextcolour}[1]{{\color{blue!80!black}#1}}
\newcommand{\targetcolour}[1]{{\color{green!40!black}#1}}
\newcommand{\altcolour}[1]{{\color{red!40!black}#1}}

\newcommand{\gacolour}[1]{{\color{Bronze}#1}}
\newcommand{\ctx}{{\boldsymbol{c}}}
\newcommand{\target}{{\boldsymbol{\alternative}^{\star}}}
\newcommand{\alternative}{{\boldsymbol{a}}}
\newcommand{\altset}[1]{\mathcal{A}_{#1}}

\newcommand{\goal}{{g}}
\newcommand{\effectiveness}{{E}}
\newcommand{\utility}{{U}}
\newcommand{\cost}{{C}}
\newcommand{\uidlocal}{\text{UID}_{\text{loc}}}
\newcommand{\uidglobal}{\text{UID}_{\text{gl}}}
\newcommand{\costsurp}{\cost_{\text{surp}}}
\newcommand{\costuidglobal}{\cost_{\uidglobal}}
\newcommand{\costuidlocal}{\cost_{\uidlocal}}
\newcommand{\costlen}{\cost_{\text{len}}}

\title{
    Surprisal Minimisation over Goal-directed Alternatives Predicts\\ Production Choice in Dialogue
}

\author{Tom Utting\aberdeen \quad Mario Giulianelli\ucl \quad Arabella Sinclair\aberdeen\ucl\\
\aberdeen University of Aberdeen \quad \ucl University College London\\
  \href{m.giulianelli@ucl.ac.uk}{\texttt{m.giulianelli@ucl.ac.uk}} \quad \href{arabella.sinclair@ucl.ac.uk}{\texttt{arabella.sinclair@ucl.ac.uk}} \\
  }

\newcommand{\ucl}[0]{$^{\triangleleft}$}
\newcommand{\aberdeen}[0]{$^{\diamond}$}

\begin{document}
\maketitle

\begin{abstract}
We model utterance production as probabilistic cost-sensitive choice over contextual alternatives, using information-theoretic notions of cost. 
We distinguish between \textit{goal-directed} alternatives that realise a fixed communicative intent and \textit{goal-agnostic} alternatives defined only by contextual plausibility, allowing us to derive speaker- and listener-oriented interpretations of different cost measures. 
We present a procedure to generate both types of alternative sets using language models. 
Analysing production choices in open-ended dialogue under both deterministic and probabilistic cost minimisation, we find that surprisal minimisation relative to goal-directed alternatives provides the strongest predictive account under both analyses.
By contrast, uniform information density and length-based costs exhibit weaker and less consistent predictive power across conditions.
More broadly, our study suggests that alternative-conditioned optimisation with LM-generated alternatives provides a principled framework for studying speaker and listener pressures in naturalistic language production.\footnote{
Code and data can be found at \url{https://github.com/the-context-lab/productionchoice}
}

\end{abstract}

\section{Introduction}

Information-theoretic and probabilistic-pragmatic models of communication provide a general framework for reasoning about utterance choice. They construe speakers as approximately rational agents that operate under resource constraints and trade off production effort against listener comprehension effort, while maintaining communicative effectiveness \cite{levy-jaeger-2006-reduction,frank2012predicting,franke2014typical,giulianelli-2022-towards,degen-rsa-framework}. 
This framing leaves open two modelling questions: 
(a) how production and comprehension costs should be defined, distinguished, and weighted against one another, and (b) over what set of alternative production choices this optimisation is assumed to take place.

This paper provides a principled way to formalise and compare speaker and listener costs in utterance production by making the space of alternatives explicit. 
Our central claim is that the interpretation of a given cost function---such as surprisal or information uniformity---depends on the set of alternative utterances with respect to which it is evaluated. 
When costs are evaluated relative to a goal-directed set of alternatives that all realise the same fixed communicative goal intended by the speaker, the cost function can be interpreted as a measure of speaker cost; correspondingly cost sensitivity yields a speaker-oriented explanation of utterance choice.
Conversely, when costs are evaluated over a goal-agnostic set of alternatives defined solely by the shared contextual state---including conversational history, common ground, and the immediate sentential context---they give rise to a listener-oriented notion of cost sensitivity.

In our experiments, we operationalise this distinction by using large language models to generate goal-directed and goal-agnostic alternative utterances.
As a case study, we identify critical choice points in production data from a conversational dialogue corpus and compute the cost of both the observed utterances and their alternatives.
We evaluate multiple cost measures---surprisal, local and global information uniformity, and length---and assess which notion of cost minimisation best accounts for speakers' observed choices.

We find that probabilistic minimisation of surprisal relative to goal-directed alternatives provides the strongest predictive account of human production choices, supporting a speaker-oriented interpretation of surprisal cost sensitivity \cite[e.g.,][]{goodman-lassiter-2015-probabilistic,futrell-2024-availability}.
These findings provide new evidence that surprisal functions as a production-side constraint in speaker decision-making, and they open avenues for studying production preferences in naturalistic language use using information-theoretic cost measures.

\section{Background}
Speakers are thought to balance their own production costs and listeners' comprehension costs when selecting among contextually available alternatives.
This section reviews prominent models of cost, with a focus on information-theoretic accounts, and examines how alternative utterances are typically handled---often only implicitly.

\subsection{Production and Comprehension Costs}
Production costs are the cognitive and temporal resources expended by speakers in formulating and realising an utterance. 
They arise from processes such as memory retrieval, advance planning of the utterance, and the impact of time pressure on formulation processes \citep{BARD2007616,Howarth01032007,Ivanova_Ferreira_2019,languages8010071}.
Comprehension costs, by contrast, reflect the cognitive and temporal effort incurred by listeners when processing and interpreting an utterance, including efforts involved in predicting upcoming words, maintaining and updating representations in working memory, resolving meaning and references, and handling rapid turn-taking
\citep{hadar2016memoryload,Peelle2017listeningeffort,meyer2023timing}.\looseness-1

Within information-theoretic models of cost, the distinction between these two categories is often blurred.
Measures such as surprisal, entropy, and uniform information density, are widely used to explain speakers' production choices \cite{genzel-charniak-2002-entropy,xu-2018-information-density,giulianelli-fernandez-2021-analysing,giulianelli-etal-2021-information,gay-etal-2026-information}, under the idea that they capture \emph{some} notion of processing cost.
However, it is often unclear whether these measures should be interpreted as proxies for comprehension cost, production cost, or a combination of the two. 
Surprisal, for example, is typically motivated as a measure of comprehension difficulty~\citep{hale-2001-probabilistic,LEVY20081126}, and has been shown to predict behavioural and neural indices of listener cost in eye-tracking, self-paced reading, brain imaging, and priming studies \citep{keller-2004-entropy,SMITH2013302,sinclair2022structural,wilcox-etal-2023-testing,jumelet2024predicting,sinclair2025priming}.
At the same time, surprisal estimates are usually derived from language models trained to approximate \emph{speaker} behaviour, and are  sometimes interpreted as reflecting speaker costs 
\citep{goodman-lassiter-2015-probabilistic,giulianelli-etal-2022-construction,yee-etal-2024-efficiency,futrell-2024-availability}.

A similar ambiguity characterises Uniform Information Density (UID) accounts of language production.
UID accounts posit that more uniform distributions of information reduce processing difficulty \citep{fenk1980konstanz,genzel-charniak-2002-entropy,turkredundancy2004,levy-jaeger-2006-reduction}, yet it is often left open whether this reduction should be attributed to speaker effort, listener effort, or properties of the communicative signal itself.
Consequently, UID has been invoked under both speaker- and listener-oriented interpretations \citep{coupe2019encoding,meister-etal-2021-revisiting,pimentel2021surprisaldurationtradeoffworldslanguages}.
This ambiguity leaves unclear how observed linguistic behaviour should be attributed to production versus comprehension costs.
In our work, we address this by modelling cost sensitivity explicitly as either speaker- or listener-oriented, evaluating utterances relative to goal-directed or goal-agnostic alternatives while preserving standard information-theoretic cost measures.

\subsection{Alternatives in Models of Production}
Computational models of language production differ substantially in the assumptions they make about the set of alternative realisations over which production and comprehension costs are evaluated. 
Many information-theoretic approaches abstract away from alternatives altogether, analysing information-theoretic properties of observed utterances or discourse without specifying the competing continuations available to the speaker \citep{genzel-charniak-2003-variation,giulianelli-fernandez-2021-analysing,tsipidi-etal-2024-surprise,tsipidi-etal-2025-harmonic}.

Other approaches, including classic Uniform Information Density accounts, implicitly assume competition among a small number of paraphrastic alternatives, such as syntactic variants that differ in how evenly information is distributed across an utterance \citep{levy-jaeger-2006-reduction,jaeger2010redundancy}.  
On the other hand, work in probabilistic pragmatics---including Rational Speech Act (RSA) models and rate–distortion approaches---typically studies optimisation over explicitly defined, highly restricted sets of alternative actions \citep{franke2014typical,goodman-lassiter-2015-probabilistic,futrell-2023-principles,futrell-2024-availability}, partly due to practical constraints on enumerating alternatives.

More recently, language models have been used to generate rich sets of contextual alternatives for investigating language comprehension and production \citep{hu-etal-2022-predicting,hu2023expectations,giulianelli-etal-2023-comes,giulianelli-etal-2023-information,giulianelli-etal-2024-generalized,giulianelli2026incremental,meister-etal-2024-towards}. This development makes it possible to operationalise probabilistic pragmatic models of production as choice models over open-ended alternative spaces. In this paper, we present methods for using language models to construct goal-directed and goal-agnostic contextual alternatives, i.e., alternatives available to the speaker and the listener, respectively.

\section{Production as Cost-sensitive Choice over Contextual Alternatives}
\label{sec:theory}
We introduce a formalisation of the language production process that enables us to distinguish between speaker- and listener-oriented costs.
Our focus is on how speakers choose among alternative continuations at a given point in an utterance, but the same formalisation applies more generally to production across different choice points and granularities, including choices between words, phrases, or clauses, as well as choices spanning sentence boundaries, given appropriate cost functions.

\subsection{Contextual Alternatives}
\label{sec:theory-alternative-sets}
Consider a classic example from \citet{jaeger2010redundancy}, which illustrates the production choice involved in realising an English complement clause:
\ex. \contextcolour{My boss confirmed} \targetcolour{that we were absolutely crazy.}

Let $\alphabet$ be a non-empty set of linguistic units (typically, though not necessarily, words) and $\strings$ the set of strings formed from units in $\alphabet$. We treat an utterance as a string\footnote{
    A string, written in boldface, is a finite sequence of units $\str = \sym_1 \ldots \sym_n$, where units are written in normal font. The length of a string is $|\str| = n$. Concatenation of strings (and units) is denoted by juxtaposition, e.g. $\str\str'$.

} and decompose it into two substrings: the \defn{context} $\ctx \in \strings$, marked in \contextcolour{blue}, and the target \defn{continuation} $\target \in \strings$, shown in \targetcolour{green}. 
In this example, the continuation is a complement clause, and the matrix verb \word{confirmed} identifies the onset of the complement clause as a decision point. We refer to such positions as \defn{choice points}: points in the production process at which the speaker selects a continuation from a set of possible contextual alternatives.

A perfectly grammatical and communicatively equivalent alternative continuation omits the complementiser \word{that}:
\ex. \contextcolour{My boss confirmed} \altcolour{we were absolutely crazy.}

Choosing this continuation $\alternative' \in \strings$ (marked in \altcolour{red}) over $\target$ constitutes a case of syntactic reduction, which leads to less uniform information distribution across the sentence and is therefore predicted to be dispreferred under UID accounts of production \cite{levy-jaeger-2006-reduction,jaeger2010redundancy}. 

The reduced and unreduced complement clauses are just two members of a much larger set of continuations that are grammatically and semantically licensed by the context, even if they differ in structure, meaning, or communicative goal. 
To characterise production choices more generally, we therefore need to consider the full space of continuations available at a given choice point.

\paragraph{Goal-agnostic alternatives.}
At a given choice point defined by context $\ctx$, the speaker is in principle free to continue the utterance in many different ways. 
We define the \defn{goal-agnostic alternative set} $\altset{\ctx}$ as the set of all continuations that are grammatically licensed and contextually coherent given $\ctx$, independently of the speaker's intended communicative goal. 
In the example above, this includes continuations such as:
\begin{itemize}
  \setlength{\itemsep}{1pt}
  \setlength{\parskip}{0pt}
  \setlength{\parsep}{0pt}
  \itemindent=-13pt
    \item[] $\alternative_{1} =$ \altcolour{we were absolutely crazy.}
    \item[] $\alternative_{2} =$ \gacolour{that the meeting has been rescheduled.}
    \item[] $\alternative_{3} =$ \gacolour{my request for time off next week.}
    \item[] $\alternative_{4} =$ \gacolour{our participation in the next conference.}
    \item[] $\alternative_{5} =$ \gacolour{that I have her full support.}
\end{itemize}
We call this alternative set goal-agnostic in that it includes continuations that would \emph{not} result in a sentence communicatively equivalent to $\ctx \target$ (marked in \gacolour{gold}). 
Formally, we assume that the goal-agnostic alternative set $\altset{\ctx}$ is obtained by sampling $N$ continuations from a distribution over strings conditioned on the context, i.e., a language model:
\begin{align}
    \altset{\ctx} \defeq \{\alternative_1, \ldots, \alternative_N\},
    \quad
    \alternative_i \sim p(\cdot \mid \ctx).
\end{align}
This alternative set reflects contextually constrained uncertainty both over which communicative goal the speaker intends to realise and over how the goal may be linguistically realised.
Accordingly, \textit{we assume that this set approximates the expectations of a \textbf{listener}}, who is uncertain about which goal the speaker will communicate.\looseness-1\footnote{
    Note that the shared contextual state (including conversational history, common ground, and immediate sentential context) can constrain the set of plausible goals.
    We therefore do not assume that the listener assigns uniform probability across all conceivable goals, but rather that they maintain uncertainty over a contextually restricted set. Accordingly, some goal-agnostic alternatives may align with the speaker’s intended goal, as is the case for $\alternative_1$ in the example above. Empirical evidence supporting this assumption is provided in \Cref{sec:goal-predictability-from-context}.\looseness-1
}

\paragraph{Goal-directed alternatives.}
By contrast, we define the \defn{goal-directed alternative set} $\altset{\ctx,\goal}$ as the set of continuations that realise a fixed communicative goal $\goal$ in context $\ctx$.
Formally, we assume that this set is obtained by sampling continuations from a distribution over strings conditioned on both the context and the communicative goal:
\begin{align}
\altset{\ctx,\goal} \defeq \{\alternative^{g}_1, \ldots, \alternative^{g}_N\},
\quad
\alternative^{g}_i \sim p(\cdot \mid \ctx, \goal).
\end{align}
Here, $p$ is not a standard language model; we show how to approximate a goal-conditioned language model in \Cref{sec:generating-alternatives}. 
In our running example, $\goal$ can be characterised informally as conveying that the speaker's judgement or behaviour was seriously mistaken. 
Examples of alternatives in $\altset{\ctx,\goal}$ include:
\begin{itemize}
  \setlength{\itemsep}{1pt}
  \setlength{\parskip}{0pt}
  \setlength{\parsep}{0pt}
  \itemindent=-13pt
    \item[] $\alternative^{g}_{1} =$ \altcolour{we were absolutely crazy.}
    \item[] $\alternative^{g}_{2} =$ \altcolour{that we were completely irrational.}
    \item[] $\alternative^{g}_{3} =$ \altcolour{that our decision made no sense at all.}
    \item[] $\alternative^{g}_{4} =$ \altcolour{that our behaviour was seriously flawed.}
    \item[] $\alternative^{g}_{5} =$ \altcolour{we were wildly off the mark.}
\end{itemize}
These alternatives vary in syntactic realisation, lexical-semantic choice, stylistic register, and evaluative
strength, but all preserve the same underlying communicative intent.
\textit{We assume that this set reflects the production uncertainty of a \textbf{speaker}}, who has fixed a communicative goal and is choosing among alternative realisations of that goal.

\subsection{A Choice Model of Production}
\label{sec:theory-choice-model}

Having defined the space of alternatives available at a choice point, we now turn to the question of how speakers choose among them.
We adopt a probabilistic, decision-theoretic perspective on production in which, in line with decision-theoretic and RSA approaches, speakers assign probability to alternative continuations in proportion to their utility.\looseness-1

Formally, let $\alternative$ denote a candidate continuation in context $\ctx$ given a fixed communicative goal $\goal$. 
The probability of the speaker producing $\alternative$ is:
\begin{align}
    P_{S}(\alternative \mid \ctx, \goal) \defpropto \exp \bigl(\alpha \, U(\alternative;\ctx,\goal)\bigr),
\end{align}
where $U(\alternative;\ctx,\goal)$ is a utility function and $\alpha \geq 0$ is a sensitivity parameter controlling the extent to which the speaker behaves as a utility-maximising agent \citep{luce1959individual}.\footnote{
    The parameter $\alpha$ is also commonly called an inverse-temperature or rationality parameter.
}
As $\alpha$ increases, the distribution becomes increasingly peaked around higher-utility utterances, converging to deterministic utility maximisation in the limit $\alpha \to \infty$; conversely, when $\alpha = 0$, choice is uniform over alternatives. 

Following standard RSA formulations, we decompose utility into an effectiveness term and a cost term:
\begin{align}
    \utility(\alternative;\ctx,\goal) \defeq \effectiveness(\alternative, \goal; \ctx) - \cost(\alternative; \ctx),
\end{align}
where $\effectiveness(\alternative, \goal; \ctx)$ denotes the \defn{communicative effectiveness} of utterance $\alternative$ for achieving goal $\goal$ in context $\ctx$, and $\cost(\alternative; \ctx)$ denotes the \defn{production cost} associated with $\alternative$ in context $\ctx$.
In much pragmatic work, production cost is held constant in order to isolate the role of communicative effectiveness. 
Here, we take the complementary perspective and isolate the role of production cost in shaping utterance choice. 
For goal-directed alternatives, this amounts to abstracting away from differences in effectiveness that are, by construction, absent, as all goal-directed continuations realise the communicative goal by definition. 
For goal-agnostic alternatives, effectiveness varies from the speaker's perspective but is uncertain from the listener's perspective, which the goal-agnostic set is intended to approximate (see \cref{sec:theory-alternative-sets}). 
In both cases, we therefore treat effectiveness as constant across alternatives.

Formally, if communicative effectiveness is held constant across alternatives---i.e.,
$\effectiveness(\alternative, \goal; \ctx) = \kappa$ for all $\alternative$ in the
relevant alternative set---then the speaker's probabilistic production rule reduces to a softmax over cost:
\begin{align}
    P_{S}(\alternative \mid \ctx, \goal)
    \defeq& \frac{\exp \bigl(\alpha(\kappa - \cost(\alternative; \ctx))\bigr)} {\sum_{\alternative' \in \altset{}} \exp \bigl(\alpha(\kappa - \cost(\alternative'; \ctx))\bigr)} \nonumber \\ 
    =\;& \frac{\exp \bigl(-\alpha \cost(\alternative; \ctx)\bigr)} {\sum_{\alternative' \in \altset{}} \exp \bigl(-\alpha \cost(\alternative'; \ctx)\bigr)} ~. \label{eq:production-rule-softmax}
\end{align}
Lower-cost alternatives are assigned higher production probability, with the strength of this preference controlled by the cost-sensitivity parameter~$\alpha$.
Assuming a finite alternative set and a real-valued cost function for which a minimum exists, this probabilistic choice rule converges, as $\alpha \to \infty$, to deterministic cost minimisation over the alternative set:\looseness-1
\begin{align}
    \target = \arg\min_{\alternative \in \altset{}} \cost(\alternative; \ctx)~. \label{eq:argmin-cost}
\end{align}
Crucially, the interpretation of this minimisation depends on the choice of alternative set $\altset{}$. 
When $\altset{} = \altset{\ctx,\goal}$, minimisation is speaker-oriented, as the alternatives differ only in how a fixed communicative goal is realised. 
When $\altset{} = \altset{\ctx}$, minimisation is listener-oriented, as the alternatives encode contextually constrained uncertainty about both the speaker’s intended goal and its realisation.
Although the speaker makes the choice in both cases, minimisation over the goal-agnostic alternative set is listener-oriented in that it reflects the speaker's model of the listener's expectations when selecting among continuations.

\section{Measures of Cost}
\label{sec:costs}
We consider four measures of utterance cost that have featured prominently in information-theoretic and probabilistic-pragmatic accounts of language production and comprehension. 

\subsection{Surprisal}
\label{sec:costs-surprisal}
Surprisal quantifies how unexpected a word or sequence is given its preceding context. For an alternative $\alternative$ in context $\ctx$, surprisal is defined as the negative log probability that a language model $p$ assigns to $\alternative$ given $\ctx$, yielding the cost:
\begin{equation}
\costsurp(\alternative; \ctx) \defeq - \log p(\alternative \mid \ctx).
\end{equation}
Within surprisal-based theories of comprehension, higher surprisal corresponds to a larger update of the comprehender's probabilistic expectations over upcoming linguistic material and thus to greater processing difficulty for the listener \citep{hale-2001-probabilistic,LEVY20081126}. Speakers are therefore predicted to prefer utterances with lower surprisal. 

At the same time, surprisal has also been interpreted as a speaker cost, for example in Rational Speech Act and rate--distortion models \citep{goodman-lassiter-2015-probabilistic,futrell-2024-availability}.
Under the rate–distortion theory of control, surprisal can be conceptualised as reflecting an ``automatic policy'' \citep{futrell-2024-availability}. Highly frequent sequences correspond to well-practised production routines that are executed relatively automatically and are therefore less costly to produce, whereas contextually unlikely continuations require suppression of this automatic policy (i.e., greater control) and are therefore more effortful to produce.

Moreover, surprisal estimates are most commonly derived from language models trained on corpora of production data. In this sense, notwithstanding their effectiveness at predicting behavioural signatures of comprehension effort, they are obtained from models that are directly optimised to approximate speaker behaviour.

\subsection{Uniform Information Density}
\label{sec:costs-uid}
A related but distinct proposal is the Uniform Information Density (UID) hypothesis, according to which speakers aim to distribute information as evenly as possible across an utterance \citep{fenk1980konstanz,turkredundancy2004,levy-jaeger-2006-reduction}. 
Avoiding sharp peaks in surprisal is argued to facilitate comprehension and to constitute a listener-oriented rational strategy given grammatical constraints. 
Following prior work \citep{collins2014information,jain-etal-2018-uniform,meister-etal-2021-revisiting}, we operationalise UID using two metrics: local and global uniformity.

\paragraph{Local Uniformity.}
We first consider \emph{local} uniformity, which quantifies the smoothness of the surprisal contour across adjacent units within a sequence. 
Given a sequence of word-level surprisals $\mathbf{s} = (s_1,\dots,s_n)$, local UID is defined as
\begin{equation} \label{eq:UID-local}
    \uidlocal(\mathbf{s}) \defeq \frac{1}{n-1} \sum_{t=2}^{n} \bigl(s_t - s_{t-1}\bigr)^2.
\end{equation}
This formulation is sensitive to fine-grained grammatical and locality-driven effects, such as the placement of optional material or function words \citep{levy-jaeger-2006-reduction}. 
In our experiments, we compute this metric only over the continuation, since surprisal values for the context are fixed across alternatives, except for a single negligible transition at the onset of the continuation. 
The corresponding cost for an alternative $\alternative$ in context $\ctx$ is\looseness-1
\begin{equation}
    \costuidlocal(\alternative; \ctx)
    \defeq
    \uidlocal\bigl(\mathbf{s}(\alternative)\bigr),
\end{equation}
where $\mathbf{s}(\alternative)$ denotes the sequence of word-level surprisals for the continuation. 
Lower values indicate more uniform and thus less costly surprisal profiles.\looseness-1

\paragraph{Global Uniformity.}
We next consider \emph{global} uniformity at the level of an entire sequence (e.g., a sentence or discourse). Given a sequence of word-level surprisals $\mathbf{s}$, global UID is defined as
\begin{equation}  \label{eq:UID-sentence}
    \uidglobal(\mathbf{s}) \defeq \frac{1}{n} \sum_{t=1}^{n} \bigl(s_t - \mu\bigr)^2,
\end{equation}
where $\mu \defeq \frac{1}{n}\sum_{t=1}^{n} s_t$ denotes the mean surprisal over the $n$ words in the sequence. 
This metric quantifies the extent to which individual surprisal values deviate from the overall sequence average; lower values indicate more uniform surprisal profiles and thus lower cost. 
Global UID has been linked to the idea of maintaining a stable average information rate over longer stretches of discourse \citep{genzel-charniak-2002-entropy,tsipidi-etal-2024-surprise}. 
In our experiments, we compute this measure over the full utterance, including both the context and the continuation, and define the corresponding cost for an alternative $\alternative$ in context $\ctx$ as
\begin{equation}
    \costuidglobal(\alternative; \ctx)
    \defeq
    \uidglobal\bigl(\mathbf{s}(\ctx \alternative)\bigr),
\end{equation}
where $\mathbf{s}(\ctx \alternative)$ denotes the sequence of word-level surprisals for the concatenation of $\ctx$ and $\alternative$.
As with local UID, global uniformity is primarily interpreted as a listener-oriented pressure. 

\paragraph{Length.}
Finally, we consider continuation length as a simple yet widely used notion of cost, where length is measured as the number of words in the continuation:
\begin{equation}
    \costlen(\alternative; \ctx)
    \defeq
    | \alternative | .
\end{equation}
Length is typically taken as a proxy for speaker effort \citep{bock1994language,degen2013cost,bergen2016pragmatic,cohn-gordon-etal-2019-incremental,white2020learning,giulianelli-2022-towards}, 
but it may also reflect pressures arising from comprehension. 

\section{Methods}
\label{sec:exp-setup}
We draw on a dialogue corpus to extract human production choices (\Cref{sec:dialouge_contexts}), estimate measures of cost using language models (\Cref{sec:estimating_costs}), and generate alternative sets following the procedure in \Cref{sec:generating-alternatives}.
Finally, we align the distributions of generated and human utterances via stratified sampling (\Cref{sec:stratified-sampling}).

\subsection{Dialogue Contexts and Continuations}
\label{sec:dialouge_contexts}
We use the Switchboard Dialogue Act Corpus \citep{stolcke-etal-2000-dialogue}, a naturalistic spoken conversational dialogue annotated with dialogue act labels.
We parse and filter the data to remove backchannels and disfluencies, which are common in spoken dialogue but are largely out of distribution for LMs. Full preprocessing details are provided in \Cref{app:methods-data-cleaning}.

We restrict the data to utterances between 10 and 30 words in length that are annotated with \textit{statement} or \textit{question} dialogue act tags and are preceded by an utterance from the other speaker. These criteria ensure that the selected utterances are complete, coherent sentences with sufficient semantic structure to serve as ground-truth human productions in our experiments. This resulted in 1,342 utterances, which we select from in \Cref{sec:generating-alternatives}. 
Each utterance is parsed and divided into a context and a continuation. 
We identify the root verb of each sentence as the choice point and define the context $\ctx$ as all material up to and including the root verb. The remaining material, from the word after the root verb to the end of the utterance, is taken to be the human continuation $\target$. This choice is motivated by Jaeger's classic that-mentioning example (see Section~\ref{sec:theory-alternative-sets} and the Limitations section for a discussion).
More details in~\Cref{app:methods-data-extract-contexts,app:methods-data-extract-targets,app:methods-data-splitting-targets}.

\subsection{Estimating Costs}
\label{sec:estimating_costs}
Three of the four cost measures we consider are surprisal-based (cf.~\Cref{sec:costs}). We estimate surprisal using GPT-2 Small \citep{radford2019language},\footnote{\href{https://huggingface.co/openai-community/gpt2}{https://huggingface.co/openai-community/gpt2}} which has been shown to align well with behavioural measures of processing effort and is widely used in psycholinguistic research \citep{oh2023why,shain2024evidence,kuribayashi-etal-2024-psychometric}. 
Surprisal of utterance contexts, as required for global UID, is computed conditional on the preceding dialogue history truncated to fit the model's context window. Surprisal of continuations is computed conditional on both the utterance context and this dialogue history.

\begin{figure*}[!t]
    \centering
    \includegraphics[width=\linewidth]{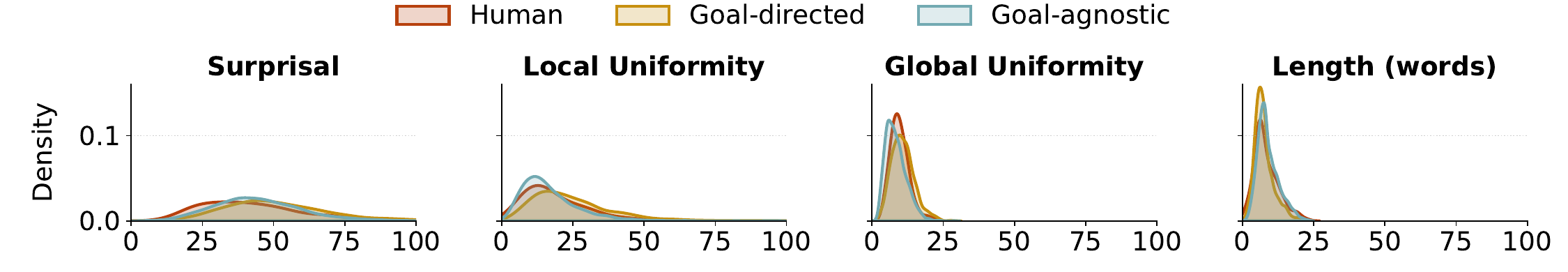}
    \caption{Global cost distribution for \textit{human}, \textit{goal-directed} and \textit{goal-agnostic} alternative sets.}
\label{fig:cost_distribution}
\vspace{-0.2em}
\end{figure*}

\subsection{Generating Alternatives}
\label{sec:generating-alternatives}

To obtain high-quality sets of alternatives for our experiments, we use an LLM as a simulator of dialogue utterance production. 
All generations were produced with OpenAI’s GPT-4o, a state-of-the-art model at the time of experimentation.\footnote{
We used the latest \href{https://platform.openai.com/docs/models/gpt-4o}{GPT-4o} snapshot available via the OpenAI API on 1 August 2025. All generations were produced with the default decoding settings, i.e., a temperature of 1 and no top-$k$ or top-$p$ truncation. All generated continuations are available at \href{https://github.com/the-context-lab/productionchoice}{github.com/the-context-lab/productionchoice}.
Full details about the generation procedure as well as additional analyses can be found in \cref{app:generating-alternatives,app:results}.
}

To generate \textbf{goal-agnostic} alternatives, we sample continuations from an LLM conditioned on different amounts of dialogue history.
We consider three history conditions: one in which the model is instructed to complete the sentence given only the utterance context; one in which it is provided with a single preceding utterance; and one in which it has access to the entire conversational history.
This reflects the fact that human speakers may vary in the extent to which they retain or rely on prior context when evaluating costs during production.
In all cases, the model is prompted to complete the sentence context via a task instruction of the form \textit{``Your task is to complete the provided sentence''}.

To generate \textbf{goal-directed} alternatives, we instruct the LLM to produce a fixed set of unique paraphrases of each observed human continuation, constrained to share the same initial context. 
We retain only those utterance contexts for which the model successfully generates at least 10 paraphrases. 
To ensure that goal-directed alternatives preserve the communicative intent of the observed utterance, we apply a post-hoc filtering and reclassification procedure to the entire dataset.
We use an LLM-as-a-judge \cite{zheng2023judging} to classify whether alternative continuations are paraphrases of the human utterance; manual annotation of 400 sampled judgements yields an accuracy of 98.75\%.
Alternatives in the goal-directed sets that are not classified as paraphrases are discarded, while goal-agnostic alternatives that are classified as paraphrases of the human utterance are also treated as goal-directed, and thus belong to both sets.
The proportion of these goal-matching goal-agnostic continuations is reported in \cref{fig:goal_match_proportions}.
This procedure yields a dataset of 12,669 items (12,360 generated and 309 observed), spanning 309 contexts.

\subsection{Aligning Human and Generated Distributions via Stratified Sampling}
\label{sec:stratified-sampling}
As outlined in \Cref{sec:theory-choice-model}, we model utterance choice as decision-making under probabilistic preferences within a given context. Under this view, for a cost measure to explain choice, human utterances are expected to exhibit lower cost than competing contextual alternatives. For this comparison to be meaningful, however, generated alternatives should not systematically differ from human utterances in their overall cost distributions, as such differences would otherwise confound context-specific effects.

We therefore assess whether, under each cost function, the distribution of generated continuations matches that of human utterances across contexts using independent-samples t-tests. 
We find no significant differences in surprisal or local uniformity, but observe small yet significant differences in length and global uniformity ($p < 0.001$).

To address this, we apply stratified sampling to the generated continuations, aligning their distributions with those of human utterances along the two affected dimensions: length and global uniformity. We first discretise human utterances into three bins per dimension and assign each utterance to a stratum defined by its $(\costlen, \costuidglobal)$ bin pair. 
The same bin boundaries are then applied to the generated continuations. 
We estimate the empirical distribution of human utterances over these strata and sample without replacement from the generated pool by selecting the largest feasible subsample whose stratum counts match the human proportions.

After stratification, differences in mean length and global UID are no longer statistically significant at the $\alpha = 0.001$ level. Figure~\ref{fig:cost_distribution} shows the resulting alignment between human and generated distributions. The final sample retains 6,335 generated utterances.\footnote{
    We obtain qualitatively identical results when using the full set of utterances; see  
    \cref{sec:app-no-stratification}.
} This adjustment ensures that subsequent analyses reflect context-specific preferences rather than global distributional differences.

\begin{figure*}[h!]
    \centering
    \includegraphics[width=\linewidth]{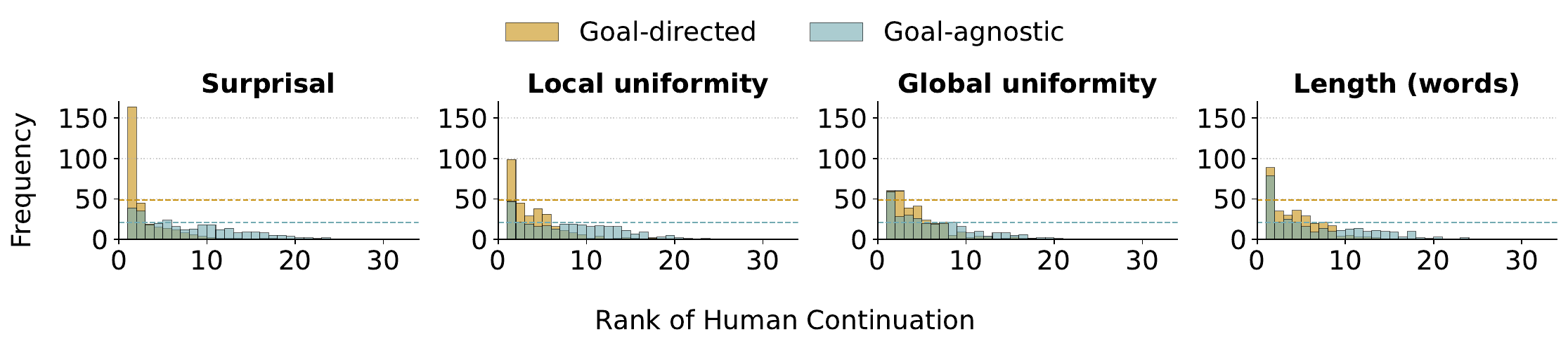}
    \caption{
    Ranking distributions of human continuations under different cost measures, evaluated against \goaldirected{goal-directed} and \goalagnostic{goal-agnostic} alternative sets. A rank of 1 indicates that the human continuation has the lowest cost amongst available alternatives, corresponding to deterministic cost minimisation.
    Dashed lines indicate chance levels.\looseness-1
    }
    \label{fig:rankingbymetric}
\end{figure*}

\section{Experiments and Results}
This section evaluates whether human production choices reflect consistent preferences for continuations that have a lower cost than their contextual alternatives, under both goal-directed and goal-agnostic alternative sets. 
In \Cref{results:cost-minimisation}, we analyse the rank of the observed continuation among its alternatives, testing whether it is the lowest-cost option more often than expected by chance. This corresponds to deterministic cost minimisation.
In \Cref{sec:condlogit}, we model production choice as probabilistic using a pairwise logistic model and directly compare the predictive strength of different cost measures.

\subsection{Deterministic Cost Minimisation}
\label{results:cost-minimisation}
To test the hypothesis that human continuations greedily minimise cost, we compute the rank of each observed continuation within both the goal-directed and goal-agnostic alternative sets. A continuation is assigned rank 1 if it minimises cost with respect to the alternative set (cf.~Eq.~\ref{eq:argmin-cost}). \Cref{fig:rankingbymetric} shows the resulting rank distributions.

To assess whether rank-1 outcomes occur more frequently than expected by chance, we use a Poisson--binomial test. 
We model the rank of the human utterance on each trial as a Bernoulli outcome indicating whether the human utterance has rank 1, with trial-specific chance levels to account for differences in the size of the alternative sets.
Statistical significance is assessed using a one-sided test.
The full specification  of the Poisson--binomial test is provided in \Cref{app:poisson–binomial}.
\Cref{tab:poisson_binomial_results} summarises the results, where the percentage of rank-1 outcomes quantifies the extent to which minimising a certain cost accounts for observed production choices.

Across all conditions, rank-1 outcomes occur significantly more often than expected by chance. The strongest absolute effect is observed for surprisal evaluated against goal-directed alternatives: 53.4\% of human continuations minimise surprisal against this set, corresponding to a 3.24$\times$ increase over the baseline. When evaluated against goal-agnostic alternatives, the percentage drops to 15.2\% (2.11$\times$). This supports a speaker-oriented interpretation of surprisal-based cost minimisation.

Utterance length shows the largest relative increase over chance in the goal-agnostic setting (3.69$\times$), exceeding all other cost measures. This suggests that length-based minimisation can be interpreted as a listener-oriented pressure to reduce processing effort.
For local and global uniformity, rank-1 outcomes are less frequent overall but still reliably above chance. 
In both cases, the relative increase over chance is higher in the goal-agnostic setting, consistent with a listener-oriented interpretation in which information content reflects uncertainty over both upcoming linguistic material and the speaker's intended goal.

Overall, surprisal shows the strongest absolute effects, with over half of continuations minimising cost under goal-directed evaluation. Length shows the strongest relative effects under goal-agnostic evaluation, with uniformity falling in between. This analysis assumes that production selects the minimum-cost alternative. We next investigate whether production choices are better described as probabilistic preferences over alternatives.\looseness-1

\begin{table}[h]
\centering
\resizebox{\linewidth}{!}{%
\begin{tabular}{l c|c}
\toprule
\textbf{Cost} & \textbf{Goal-directed} & \textbf{Goal-agnostic} \\
\midrule
Surprisal & 53.4\% {\scriptsize\textcolor{YlOrBr8}{$\bf\times 3.24$}} & 15.2\% {\scriptsize\textcolor{YlOrBr5}{$\bf\times 2.11$}} \\
Local uniformity & 34.1\% {\scriptsize\textcolor{YlOrBr5}{$\bf\times 2.07$}} & 16.2\% {\scriptsize\textcolor{YlOrBr6}{$\bf\times 2.25$}} \\
Global uniformity & 24.1\% {\scriptsize\textcolor{YlOrBr3}{$\bf\times 1.46$}} & 19.3\% {\scriptsize\textcolor{YlOrBr7}{$\bf\times 2.68$}} \\
Length (words) & 28.6\% {\scriptsize\textcolor{YlOrBr4}{$\bf\times 1.73$}} & 26.6\% {\scriptsize\textcolor{YlOrBr9}{$\bf\times 3.69$}} \\
\midrule
Uniform \textit{(baseline)}  & 16.5\% & 7.2\% \\
\bottomrule
\end{tabular}%
}
\caption{Percentage of observed rank-1 outcomes, with multiplicative increase relative to the uniform baseline (darker indicates larger deviation from the baseline). All values are significantly higher than chance under one-sided Poisson--binomial tests ($p < 10^{-5}$ or smaller).\looseness-1}
\label{tab:poisson_binomial_results}
\end{table}

\begin{figure*}[t]
    \centering
    \includegraphics[width=\textwidth]{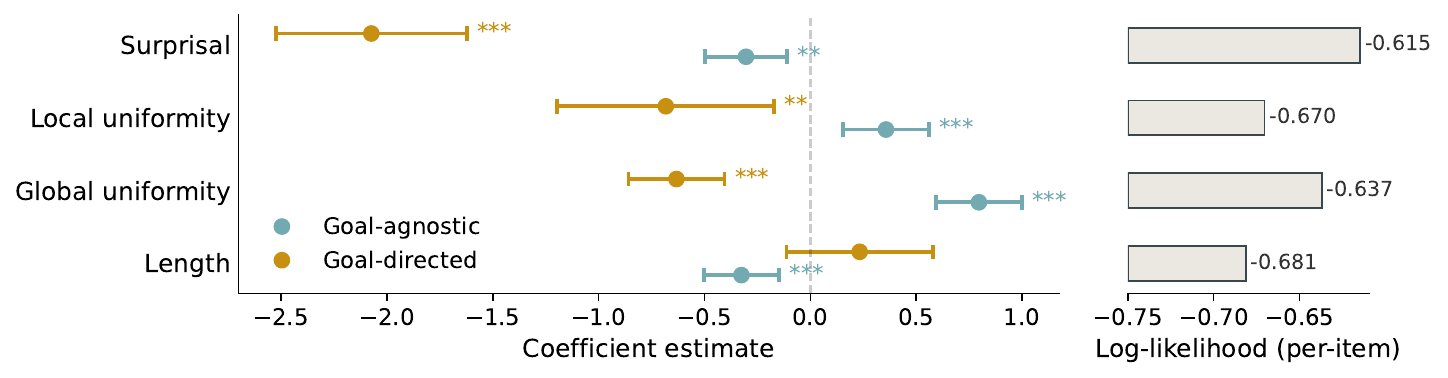}
    \caption{
    Logistic regression results predicting whether a continuation is selected over an alternative as a function of the difference in cost with respect to the alternative, goal condition, and their interaction. Points show coefficient estimates on the log-odds scale, horizontal bars indicate 95\% confidence intervals. Asterisks indicate significance levels ($p < .05$, $p < .01$, $p < .001$). On the right panel is the corresponding per-item log-likelihood for each model.
    }
    \label{fig:logit_coefficients_loglik}
\end{figure*}

\subsection{Graded Cost Sensitivity}
\label{sec:condlogit}
The rank-based analysis in \Cref{results:cost-minimisation} corresponds to a limiting case of the probabilistic choice model in which choice noise vanishes and the speaker deterministically selects the cost-minimising alternative (cf.~\Cref{sec:theory-choice-model}). 
We now relax this assumption and instead model production choice as probabilistic using a using a pairwise logistic choice model. 

We recast each observation as a binary comparison between the human continuation and a contextual alternative. The probability that a continuation $\alternative_i$ is preferred to an alternative $\alternative_j$ is modelled as a logistic function of their cost difference:
\begin{align}
P(\alternative_i \succ \alternative_j; \ctx)
= \sigma\big(\alpha (\cost(\alternative_j; \ctx) - \cost(\alternative_i; \ctx))\big) 
\end{align}
This corresponds to a two-alternative reduction of the Luce-style choice rule introduced in \Cref{sec:theory-choice-model}. We adopt a pairwise formulation because the goal-directed and goal-agnostic alternative sets differ both in composition and size, which makes coefficients from standard discrete-choice models, such as conditional logit, not directly comparable across the two conditions.\footnote{
    For completeness, we report conditional logit models fitted separately for goal-directed and goal-agnostic alternatives in \Cref{app:conditional_logit}. These yield qualitatively similar patterns but do not permit direct cross-condition comparison.
} 
To test whether cost sensitivity differs between the goal-directed and the goal-agnostic condition, we include an interaction between cost differences and the goal condition. 
Standard errors are clustered at the context level to account for the non-independence induced by multiple comparisons within the same context. Full model details are provided in \Cref{app:pairwise-logit}.

\Cref{fig:logit_coefficients_loglik} summarises the results. The surprisal-based model achieves the highest log-likelihood. Surprisal exhibits a consistent negative effect, with lower surprisal increasing the probability of selection. This effect is substantially stronger (approx.~7$\times$) in the goal-directed condition than in the goal-agnostic condition, supporting the interpretation of surprisal as a speaker-oriented cost: among continuations that realise the same goal, speakers preferentially select those with lower surprisal.

For local and global uniformity, lower cost (i.e., a more uniform information profile) predicts choice in the goal-directed condition, 
but this relationship reverses in the goal-agnostic condition, where higher cost (lower uniformity) is associated with higher choice probability. 
In conjunction with \Cref{results:cost-minimisation}, this suggests that although human continuations minimise uniformity-based costs at above-chance rates, this does not extend to a general probabilistic preference for higher-uniformity continuations.\looseness-1

Finally, we find a preference for shorter continuations in the goal-agnostic condition, 
but no effect in the goal-directed condition. In combination with the rank-based results, this suggests that length operates as a listener-oriented pressure, though the model itself yields the weakest overall fit.

\section{Conclusion}
In this paper, we argued that notions of production cost must be interpreted relative to the alternative utterances over which they are evaluated.
By explicitly distinguishing between goal-directed and goal-agnostic alternative sets, we showed that the same cost measure can give rise to qualitatively different interpretations. 
Our empirical analyses indicate that surprisal minimisation over goal-directed alternatives provides the strongest account of production choices at the main-verb choice point. This effect is robust across both rank-based and probabilistic analyses and reflects a speaker-oriented pressure to select low-surprisal continuations among alternatives that realise the same communicative goal.

Methodologically, we introduced scalable procedures for constructing contextual alternative sets that enable probabilistic pragmatic models of production to be instantiated and evaluated at scale in open-ended settings, while providing a principled basis for comparing competing notions of cost. 
More broadly, our results highlight the importance of making alternative spaces explicit when interpreting information-theoretic measures of cost, and suggest that alternative-conditioned optimisation offers a fruitful framework for studying speaker and listener pressures in naturalistic language use and for reproducing them in generation systems. 

\section*{Limitations}
Firstly, while the theoretical and methodological approach forms a core part of our contributions, the generalisability of our empirical findings is limited to the English language dataset we choose as our case study, and the relatively small subset of samples we select. 
It remains for future work to assess the extent to which these results generalise to other dialogue corpora, production types, and languages.

Second, our analysis focuses on a single class of choice points, matrix verb continuations. Although this provides a controlled setting and directly compares to classic work on production choice, it limits the scope of the empirical claims. Future work could apply the same framework to a broader range of syntactically and information-theoretically determined choice points, such as dative alternations or positions of high continuation entropy, in order to assess the generality of the observed patterns.

Third, our cost metrics are computed as global aggregates over entire continuations. While we control for length by selecting relatively short human continuations and matching the distributions of generated alternatives, this approach may become problematic for longer utterances, where contributions from later parts of the string can reduce sensitivity to the region immediately following the choice-point. More refined aggregation schemes---such as weighting units by their distance from the choice point---may better capture incremental planning and more realistic production horizons.

Fourth, our analyses rely on language models both to estimate information-theoretic quantities and to generate alternative sets. 
These models may not fully capture human processing or behaviour and may be ill-suited for simulating alternatives in certain contexts, 
especially settings that are poorly represented in training data, including low-resource languages and under-represented speaker communities within a language.

Finally, our framework does not incorporate an explicit notion of communicative effectiveness. In the open-ended dialogue setting we consider, there is no clear external success signal, and well-formed utterances between competent speakers can be assumed to be broadly understood, thus making effectiveness effectively constant across alternatives. 
Introducing a more nuanced notion of effectiveness would require an explicit model of listener interpretation, which remains an open challenge and lies beyond the scope of the present work.

\section*{Acknowledgments}
We thank the anonymous ARR reviewers for their helpful comments. We disclose the use of generative AI tools for light editing and rephrasing; the original text was our own, and we carefully reviewed all suggested edits.

\bibliography{custom}

@inproceedings{genzel-charniak-2002-entropy, 
    title={Entropy Rate Constancy in Text},
    author={Genzel, Dmitriy  and Charniak, Eugene},
    booktitle={Proceedings of the 40th Annual Meeting of the Association for Computational Linguistics},
    month={jul},
    year={2002},
    address={Philadelphia, Pennsylvania, USA},
    publisher={Association for Computational Linguistics},
    url={https://aclanthology.org/P02-1026},
    doi={10.3115/1073083.1073117},
    pages={199--206},
}

@inproceedings{genzel-charniak-2003-variation, 
    title={Variation of Entropy and Parse Trees of Sentences as a Function of the Sentence Number},
    author={Genzel, Dmitriy  and Charniak, Eugene},
    booktitle={Proceedings of the 2003 Conference on Empirical Methods in Natural Language Processing},
    year={2003},
    url={https://aclanthology.org/W03-1009},
    pages={65--72},
}

@inproceedings{levy-jaeger-2006-reduction,
 author = {Levy, Roger and Jaeger, T. Florian},
 booktitle = {Advances in Neural Information Processing Systems},
 editor = {B. Sch\"{o}lkopf and J. Platt and T. Hoffman},
 pages = {},
 publisher = {MIT Press},
 title = {Speakers optimize information density through syntactic reduction},
 url = {https://proceedings.neurips.cc/paper_files/paper/2006/file/c6a01432c8138d46ba39957a8250e027-Paper.pdf},
 volume = {19},
 year = {2006}
}

@article{fenk1980konstanz, 
    title={Konstanz im {K}urzzeitged{\"a}chtnis --{K}onstanz im sprachlichen {I}nformationsflu{\ss}?},
    author={Fenk, August and Fenk, Gertraud},
    journal={Zeitschrift f{\"u}r experimentelle und angewandte Psychologie},
    volume={27},
    number={3},
    pages={400--414},
    year={1980},
    url={http://wwwg.uni-klu.ac.at/mk0/personal/bedienst/Kurzzeitgedaechtnis1980.pdf}, 
}

@inproceedings{keller-2004-entropy, 
    title={The Entropy Rate Principle as a Predictor of Processing Effort: {A}n Evaluation against Eye-tracking Data},
    author={Keller, Frank},
    editor={Lin, Dekang  and Wu, Dekai},
    booktitle={Proceedings of the 2004 Conference on Empirical Methods in Natural Language Processing},
    month={jul},
    year={2004},
    address={Barcelona, Spain},
    publisher={Association for Computational Linguistics},
    url={https://aclanthology.org/W04-3241},
    pages={317--324},
}

@inproceedings{giulianelli-fernandez-2021-analysing, 
    title={Analysing Human Strategies of Information Transmission as a Function of Discourse Context},
    author={Giulianelli, Mario  and Fern{\'a}ndez, Raquel},
    editor={Bisazza, Arianna  and Abend, Omri},
    booktitle={Proceedings of the 25th Conference on Computational Natural Language Learning},
    month={nov},
    year={2021},
    address={Online},
    publisher={Association for Computational Linguistics},
    url={https://aclanthology.org/2021.conll-1.50},
    doi={10.18653/v1/2021.conll-1.50},
    pages={647--660},
}

@inproceedings{meister-etal-2021-revisiting, 
    title={Revisiting the {U}niform {I}nformation {D}ensity Hypothesis},
    author={Meister, Clara  and Pimentel, Tiago  and Haller, Patrick  and J{\"a}ger, Lena  and Cotterell, Ryan  and Levy, Roger},
    editor={Moens, Marie-Francine  and Huang, Xuanjing  and Specia, Lucia  and Yih, Scott Wen-tau},
    booktitle={Proceedings of the 2021 Conference on Empirical Methods in Natural Language Processing},
    month={nov},
    year={2021},
    address={Online and Punta Cana, Dominican Republic},
    publisher={Association for Computational Linguistics},
    url={https://aclanthology.org/2021.emnlp-main.74},
    doi={10.18653/v1/2021.emnlp-main.74},
    pages={963--980},
}

@inproceedings{giulianelli-etal-2021-information, 
    title={Is Information Density Uniform in Task-Oriented Dialogues?},
    author={Giulianelli, Mario  and Sinclair, Arabella  and Fern{\'a}ndez, Raquel},
    editor={Moens, Marie-Francine  and Huang, Xuanjing  and Specia, Lucia  and Yih, Scott Wen-tau},
    booktitle={Proceedings of the 2021 Conference on Empirical Methods in Natural Language Processing},
    month={nov},
    year={2021},
    address={Online and Punta Cana, Dominican Republic},
    publisher={Association for Computational Linguistics},
    url={https://aclanthology.org/2021.emnlp-main.652},
    doi={10.18653/v1/2021.emnlp-main.652},
    pages={8271--8283},
}

@article{jaeger2010redundancy, 
    author={Jaeger, T. Florian},
    journal={Cognitive Psychology},
    number={1},
    pages={23--62},
    title={Redundancy and reduction: {S}peakers manage syntactic information density},
    volume={61},
    year={2010},
    url={https://www.sciencedirect.com/science/article/pii/S0010028510000083}, 
}

@article{collins2014information, 
    title={Information density and dependency length as complementary cognitive models},
    author={Collins, Michael Xavier},
    journal={Journal of psycholinguistic research},
    volume={43},
    pages={651--681},
    year={2014},
    url={https://idp.springer.com/authorize/casa?redirect_uri=https://link.springer.com/article/10.1007/s10936-013-9273-3}, 
}

@article{frank2012predicting,
  title={Predicting pragmatic reasoning in language games},
  author={Frank, Michael C and Goodman, Noah D},
  journal={Science},
  volume={336},
  number={6084},
  pages={998--998},
  year={2012},
  publisher={American Association for the Advancement of Science}
}

@inproceedings{giulianelli-2022-towards,
    title = "Towards Pragmatic Production Strategies for Natural Language Generation Tasks",
    author = "Giulianelli, Mario",
    editor = "Goldberg, Yoav  and
      Kozareva, Zornitsa  and
      Zhang, Yue",
    booktitle = "Proceedings of the 2022 Conference on Empirical Methods in Natural Language Processing",
    month = dec,
    year = "2022",
    address = "Abu Dhabi, United Arab Emirates",
    publisher = "Association for Computational Linguistics",
    url = "https://aclanthology.org/2022.emnlp-main.544",
    doi = "10.18653/v1/2022.emnlp-main.544",
    pages = "7978--7984"
}

@article{
futrell-2023-principles,
author = {Richard Futrell },
title = {Information-theoretic principles in incremental language production},
journal = {Proceedings of the National Academy of Sciences},
volume = {120},
number = {39},
pages = {e2220593120},
year = {2023},
doi = {10.1073/pnas.2220593120},
URL = {https://www.pnas.org/doi/abs/10.1073/pnas.2220593120},
eprint = {https://www.pnas.org/doi/pdf/10.1073/pnas.2220593120}}

@article{futrell-2024-availability,
author = {Futrell, Richard},
title = {An Information-Theoretic Account of Availability Effects in Language Production},
journal = {Topics in Cognitive Science},
volume = {16},
number = {1},
pages = {38-53},
doi = {https://doi.org/10.1111/tops.12716},
url = {https://onlinelibrary.wiley.com/doi/abs/10.1111/tops.12716},
eprint = {https://onlinelibrary.wiley.com/doi/pdf/10.1111/tops.12716},
year = {2024}
}

@inproceedings{giulianelli-etal-2022-construction,
    title = "Construction Repetition Reduces Information Rate in Dialogue",
    author = "Giulianelli, Mario  and
      Sinclair, Arabella  and
      Fern{\'a}ndez, Raquel",
    editor = "He, Yulan  and
      Ji, Heng  and
      Li, Sujian  and
      Liu, Yang  and
      Chang, Chua-Hui",
    booktitle = "Proceedings of the 2nd Conference of the Asia-Pacific Chapter of the Association for Computational Linguistics and the 12th International Joint Conference on Natural Language Processing (Volume 1: Long Papers)",
    month = nov,
    year = "2022",
    address = "Online only",
    publisher = "Association for Computational Linguistics",
    url = "https://aclanthology.org/2022.aacl-main.51/",
    doi = "10.18653/v1/2022.aacl-main.51",
    pages = "665--682"
}

@inproceedings{yee-etal-2024-efficiency,
    title = "Efficiency and Effectiveness in Task-Oriented Dialogue: On Construction Repetition, Information Rate, and Task Success",
    author = "Yee, Jun Sen  and
      Giulianelli, Mario  and
      Sinclair, Arabella J.",
    editor = "Calzolari, Nicoletta  and
      Kan, Min-Yen  and
      Hoste, Veronique  and
      Lenci, Alessandro  and
      Sakti, Sakriani  and
      Xue, Nianwen",
    booktitle = "Proceedings of the 2024 Joint International Conference on Computational Linguistics, Language Resources and Evaluation (LREC-COLING 2024)",
    month = may,
    year = "2024",
    address = "Torino, Italia",
    publisher = "ELRA and ICCL",
    url = "https://aclanthology.org/2024.lrec-main.494/",
    pages = "5562--5577"
}

@inproceedings{hale-2001-probabilistic,
    title = "A Probabilistic {E}arley Parser as a Psycholinguistic Model",
    author = "Hale, John",
    booktitle = "Second Meeting of the North {A}merican Chapter of the Association for Computational Linguistics",
    year = "2001",
    url = "https://aclanthology.org/N01-1021/"
}

@article{SMITH2013302,
title = {The effect of word predictability on reading time is logarithmic},
journal = {Cognition},
volume = {128},
number = {3},
pages = {302-319},
year = {2013},
issn = {0010-0277},
doi = {https://doi.org/10.1016/j.cognition.2013.02.013},
url = {https://www.sciencedirect.com/science/article/pii/S0010027713000413},
author = {Nathaniel J. Smith and Roger Levy}
}

@article{LEVY20081126,
title = {Expectation-based syntactic comprehension},
journal = {Cognition},
volume = {106},
number = {3},
pages = {1126-1177},
year = {2008},
issn = {0010-0277},
doi = {https://doi.org/10.1016/j.cognition.2007.05.006},
url = {https://www.sciencedirect.com/science/article/pii/S0010027707001436},
author = {Roger Levy}
}

@article{sinclair2022structural,
   title = "Structural Persistence in Language Models: Priming as a Window into Abstract Language Representations",
    author = "Sinclair, Arabella  and
      Jumelet, Jaap  and
      Zuidema, Willem  and
      Fern{\'a}ndez, Raquel",
    editor = "Roark, Brian  and
      Nenkova, Ani",
    journal = "Transactions of the Association for Computational Linguistics",
    volume = "10",
    year = "2022",
    address = "Cambridge, MA",
    publisher = "MIT Press",
    url = "https://aclanthology.org/2022.tacl-1.60/",
    doi = "10.1162/tacl_a_00504",
    pages = "1031--1050"
}

@article{jumelet2024predicting,
    title = "Do Language Models Exhibit Human-like Structural Priming Effects?",
    author = "Jumelet, Jaap  and
      Zuidema, Willem  and
      Sinclair, Arabella",
    editor = "Ku, Lun-Wei  and
      Martins, Andre  and
      Srikumar, Vivek",
    journal = "Findings of the Association for Computational Linguistics: ACL 2024",
    month = aug,
    year = "2024",
    address = "Bangkok, Thailand",
    publisher = "Association for Computational Linguistics",
    url = "https://aclanthology.org/2024.findings-acl.877/",
    doi = "10.18653/v1/2024.findings-acl.877",
    pages = "14727--14742"
}

@article{sinclair2025priming,
title = {Structural priming in humans and large language models},
journal = {Journal of Memory and Language},
volume = {149},
pages = {104713},
year = {2026},
issn = {0749-596X},
doi = {https://doi.org/10.1016/j.jml.2025.104713},
url = {https://www.sciencedirect.com/science/article/pii/S0749596X25001068},
author = {Arabella Sinclair and Anastasia Klimovich-Gray and Jaap Jumelet and Nika Adamian and Agnieszka Konopka},
}

@inproceedings{molnar2023attribution,
    title = "Attribution and Alignment: Effects of Local Context Repetition on Utterance Production and Comprehension in Dialogue",
    author = "Molnar, Aron  and
      Jumelet, Jaap  and
      Giulianelli, Mario  and
      Sinclair, Arabella",
    editor = "Jiang, Jing  and
      Reitter, David  and
      Deng, Shumin",
    booktitle = "Proceedings of the 27th Conference on Computational Natural Language Learning (CoNLL)",
    month = dec,
    year = "2023",
    address = "Singapore",
    publisher = "Association for Computational Linguistics",
    url = "https://aclanthology.org/2023.conll-1.18/",
    doi = "10.18653/v1/2023.conll-1.18",
    pages = "254--273"
}

@inproceedings{
zheng2023judging,
title={Judging {LLM}-as-a-Judge with {MT}-{B}ench and {C}hatbot {A}rena},
author={Lianmin Zheng and Wei-Lin Chiang and Ying Sheng and Siyuan Zhuang and Zhanghao Wu and Yonghao Zhuang and Zi Lin and Zhuohan Li and Dacheng Li and Eric Xing and Hao Zhang and Joseph E. Gonzalez and Ion Stoica},
booktitle={Thirty-seventh Conference on Neural Information Processing Systems Datasets and Benchmarks Track},
year={2023},
url={https://openreview.net/forum?id=uccHPGDlao}
}

@incollection{bock1994language,
  title={Language production: Grammatical encoding},
  author={Bock, Kathryn and Levelt, Willem JM},
  booktitle={Handbook of psycholinguistics},
  pages={945--984},
  year={1994},
  publisher={Academic Press}
}

@article{turkredundancy2004,
author = {Matthew Aylett and Alice Turk},
title ={The Smooth Signal Redundancy Hypothesis: A Functional Explanation for Relationships between Redundancy, Prosodic Prominence, and Duration in Spontaneous Speech},

journal = {Language and Speech},
volume = {47},
number = {1},
pages = {31-56},
year = {2004},
doi = {10.1177/00238309040470010201},
    note ={PMID: 15298329},

URL = { 
    
        https://doi.org/10.1177/00238309040470010201

},
eprint = { 
    
        https://doi.org/10.1177/00238309040470010201

}
}

@article{coupe2019encoding,
author = {Christophe Coupé  and Yoon Mi Oh  and Dan Dediu  and François Pellegrino },
title = {Different languages, similar encoding efficiency: Comparable information rates across the human communicative niche},
journal = {Science Advances},
volume = {5},
number = {9},
pages = {eaaw2594},
year = {2019},
doi = {10.1126/sciadv.aaw2594},
URL = {https://www.science.org/doi/abs/10.1126/sciadv.aaw2594},
eprint = {https://www.science.org/doi/pdf/10.1126/sciadv.aaw2594}
}

@misc{pimentel2021surprisaldurationtradeoffworldslanguages,
      title={A surprisal--duration trade-off across and within the world's languages}, 
      author={Tiago Pimentel and Clara Meister and Elizabeth Salesky and Simone Teufel and Damián Blasi and Ryan Cotterell},
      year={2021},
      eprint={2109.15000},
      archivePrefix={arXiv},
      primaryClass={cs.CL},
      url={https://arxiv.org/abs/2109.15000}, 
}

@inproceedings{hu-etal-2022-predicting,
    title = "Predicting scalar diversity with context-driven uncertainty over alternatives",
    author = "Hu, Jennifer  and
      Levy, Roger  and
      Schuster, Sebastian",
    editor = "Chersoni, Emmanuele  and
      Hollenstein, Nora  and
      Jacobs, Cassandra  and
      Oseki, Yohei  and
      Pr{\'e}vot, Laurent  and
      Santus, Enrico",
    booktitle = "Proceedings of the Workshop on Cognitive Modeling and Computational Linguistics",
    month = may,
    year = "2022",
    address = "Dublin, Ireland",
    publisher = "Association for Computational Linguistics",
    url = "https://aclanthology.org/2022.cmcl-1.8/",
    doi = "10.18653/v1/2022.cmcl-1.8",
    pages = "68--74"
}

@Article{languages8010071,
AUTHOR = {Betz, Simon and Bryhadyr, Nataliya and Türk, Olcay and Wagner, Petra},
TITLE = {Cognitive Load Increases Spoken and Gestural Hesitation Frequency},
JOURNAL = {Languages},
VOLUME = {8},
YEAR = {2023},
NUMBER = {1},
ARTICLE-NUMBER = {71},
URL = {https://www.mdpi.com/2226-471X/8/1/71},
ISSN = {2226-471X},
DOI = {10.3390/languages8010071}
}

@article{Ivanova_Ferreira_2019, title={The role of working memory for syntactic formulation in language production.}, volume={45}, DOI={10.1037/xlm0000672}, number={10}, journal={Journal of Experimental Psychology: Learning, Memory, and Cognition}, author={Ivanova, Iva and Ferreira, Victor S.}, year={2019}, month={Oct}, pages={1791–1814}}

@article{BARD2007616,
title = {Let’s you do that: Sharing the cognitive burdens of dialogue},
journal = {Journal of Memory and Language},
volume = {57},
number = {4},
pages = {616-641},
year = {2007},
note = {Language-Vision Interaction},
issn = {0749-596X},
doi = {https://doi.org/10.1016/j.jml.2006.12.003},
url = {https://www.sciencedirect.com/science/article/pii/S0749596X07000113},
author = {E.G. Bard and A.H. Anderson and Y. Chen and H.B.M. Nicholson and C. Havard and S. Dalzel-Job}
}

@article{Howarth01032007,
author = {Barbara Howarth and Anne H. Anderson},
title = {Introducing objects in spoken dialogue: The influence of conversational setting and cognitive load on the articulation and use of referring expressions},
journal = {Language and Cognitive Processes},
volume = {22},
number = {2},
pages = {272--296},
year = {2007},
publisher = {Routledge},
doi = {10.1080/01690960600632796},
URL = { 
    
        https://doi.org/10.1080/01690960600632796
},
eprint = { 
        https://doi.org/10.1080/01690960600632796
}

}

@article{hadar2016memoryload,
author = {Hadar, Britt and Skrzypek, Joshua and Wingfield, Arthur and Ben-David, Boaz},
year = {2016},
month = {05},
pages = {},
title = {Working Memory Load Affects Processing Time in Spoken Word Recognition: Evidence from Eye-Movements},
volume = {10},
journal = {Frontiers in Neuroscience},
doi = {10.3389/fnins.2016.00221}
}

@article{meyer2023timing,
author = {Meyer, Antje},
year = {2023},
month = {04},
pages = {},
title = {Timing in Conversation},
volume = {6},
journal = {Journal of Cognition},
doi = {10.5334/joc.268}
}

@article{Peelle2017listeningeffort,
author = {Peelle, Jonathan},
year = {2017},
month = {09},
pages = {1},
title = {Listening Effort: How the Cognitive Consequences of Acoustic Challenge Are Reflected in Brain and Behavior},
volume = {39},
journal = {Ear and Hearing},
doi = {10.1097/AUD.0000000000000494}
}

@article{degen-rsa-framework,
   author = "Degen, Judith",
   title = "The Rational Speech Act Framework", 
   journal= "Annual Review of Linguistics",
   year = "2023",
   volume = "9",
   number = "Volume 9, 2023",
   pages = "519-540",
   doi = "https://doi.org/10.1146/annurev-linguistics-031220-010811",
   url = "https://www.annualreviews.org/content/journals/10.1146/annurev-linguistics-031220-010811",
   publisher = "Annual Reviews",
   issn = "2333-9691",
   type = "Journal Article",
  }

@article{xu-2018-information-density,
title = {Information density converges in dialogue: Towards an information-theoretic model},
journal = {Cognition},
volume = {170},
pages = {147-163},
year = {2018},
issn = {0010-0277},
doi = {https://doi.org/10.1016/j.cognition.2017.09.018},
url = {https://www.sciencedirect.com/science/article/pii/S0010027717302615},
author = {Yang Xu and David Reitter}
}

@article{stolcke-etal-2000-dialogue,
    title = "Dialogue act modeling for automatic tagging and recognition of conversational speech",
    author = "Stolcke, Andreas  and
      Ries, Klaus  and
      Coccaro, Noah  and
      Shriberg, Elizabeth  and
      Bates, Rebecca  and
      Jurafsky, Daniel  and
      Taylor, Paul  and
      Martin, Rachel  and
      Van Ess-Dykema, Carol  and
      Meteer, Marie",
    journal = "Computational Linguistics",
    volume = "26",
    number = "3",
    year = "2000",
    address = "Cambridge, MA",
    publisher = "MIT Press",
    url = "https://aclanthology.org/J00-3003/",
    pages = "339--374"
}

@article{radford2019language,
  author = {Radford, Alec and Wu, Jeffrey and Child, Rewon and Luan, David and Amodei, Dario and Sutskever, Ilya},
  journal = {OpenAI},
  note = {Accessed: 2024-11-15},
  title = {Language Models are Unsupervised Multitask Learners},
  url = {https://cdn.openai.com/better-language-models/language_models_are_unsupervised_multitask_learners.pdf},
  year = 2019
}

@misc{williams2018broadcoveragechallengecorpussentence,
      title={A Broad-Coverage Challenge Corpus for Sentence Understanding through Inference}, 
      author={Adina Williams and Nikita Nangia and Samuel R. Bowman},
      year={2018},
      eprint={1704.05426},
      archivePrefix={arXiv},
      primaryClass={cs.CL},
      url={https://arxiv.org/abs/1704.05426}, 
}

@misc{liu2019robertarobustlyoptimizedbert,
      title={RoBERTa: A Robustly Optimized BERT Pretraining Approach}, 
      author={Yinhan Liu and Myle Ott and Naman Goyal and Jingfei Du and Mandar Joshi and Danqi Chen and Omer Levy and Mike Lewis and Luke Zettlemoyer and Veselin Stoyanov},
      year={2019},
      eprint={1907.11692},
      archivePrefix={arXiv},
      primaryClass={cs.CL},
      url={https://arxiv.org/abs/1907.11692}, 
}

@misc{chung2022scalinginstructionfinetunedlanguagemodels,
      title={Scaling Instruction-Finetuned Language Models}, 
      author={Hyung Won Chung and Le Hou and Shayne Longpre and Barret Zoph and Yi Tay and William Fedus and Yunxuan Li and Xuezhi Wang and Mostafa Dehghani and Siddhartha Brahma and Albert Webson and Shixiang Shane Gu and Zhuyun Dai and Mirac Suzgun and Xinyun Chen and Aakanksha Chowdhery and Alex Castro-Ros and Marie Pellat and Kevin Robinson and Dasha Valter and Sharan Narang and Gaurav Mishra and Adams Yu and Vincent Zhao and Yanping Huang and Andrew Dai and Hongkun Yu and Slav Petrov and Ed H. Chi and Jeff Dean and Jacob Devlin and Adam Roberts and Denny Zhou and Quoc V. Le and Jason Wei},
      year={2022},
      eprint={2210.11416},
      archivePrefix={arXiv},
      primaryClass={cs.LG},
      url={https://arxiv.org/abs/2210.11416}, 
}

@misc{mirzadeh2025gsmsymbolicunderstandinglimitationsmathematical,
      title={GSM-Symbolic: Understanding the Limitations of Mathematical Reasoning in Large Language Models}, 
      author={Iman Mirzadeh and Keivan Alizadeh and Hooman Shahrokhi and Oncel Tuzel and Samy Bengio and Mehrdad Farajtabar},
      year={2025},
      eprint={2410.05229},
      archivePrefix={arXiv},
      primaryClass={cs.LG},
      url={https://arxiv.org/abs/2410.05229}, 
}

@inproceedings{giulianelli-etal-2023-information,
    title = "Information Value: {M}easuring Utterance Predictability as Distance from Plausible Alternatives",
    author = "Giulianelli, Mario  and
      Wallbridge, Sarenne  and
      Fern{\'a}ndez, Raquel",
    editor = "Bouamor, Houda  and
      Pino, Juan  and
      Bali, Kalika",
    booktitle = "Proceedings of the 2023 Conference on Empirical Methods in Natural Language Processing",
    month = dec,
    year = "2023",
    address = "Singapore",
    publisher = "Association for Computational Linguistics",
    url = "https://aclanthology.org/2023.emnlp-main.343",
    doi = "10.18653/v1/2023.emnlp-main.343",
    pages = "5633--5653",
}

@incollection{goodman-lassiter-2015-probabilistic,
author = {Goodman, Noah D. and Lassiter, Daniel},
publisher = {John Wiley \& Sons, Ltd},
isbn = {9781118882139},
title = {Probabilistic Semantics and Pragmatics Uncertainty in Language and Thought},
booktitle = {The Handbook of Contemporary Semantic Theory},
chapter = {21},
pages = {655-686},
doi = {https://doi.org/10.1002/9781118882139.ch21},
url = {https://onlinelibrary.wiley.com/doi/abs/10.1002/9781118882139.ch21},
eprint = {https://onlinelibrary.wiley.com/doi/pdf/10.1002/9781118882139.ch21},
year = {2015}
}

@book{luce1959individual,
  title={Individual choice behavior},
  author={Luce, R Duncan},
  volume={4},
  year={1959},
  publisher={Wiley New York}
}

@inproceedings{jain-etal-2018-uniform,
    title = "{U}niform {I}nformation {D}ensity Effects on Syntactic Choice in {H}indi",
    author = "Jain, Ayush  and
      Singh, Vishal  and
      Ranjan, Sidharth  and
      Rajkumar, Rajakrishnan  and
      Agarwal, Sumeet",
    editor = "Becerra-Bonache, Leonor  and
      Jim{\'e}nez-L{\'o}pez, M. Dolores  and
      Mart{\'i}n-Vide, Carlos  and
      Torrens-Urrutia, Adri{\`a}",
    booktitle = "Proceedings of the Workshop on Linguistic Complexity and Natural Language Processing",
    month = aug,
    year = "2018",
    address = "Santa Fe, New-Mexico",
    publisher = "Association for Computational Linguistics",
    url = "https://aclanthology.org/W18-4605/",
    pages = "38--48"
}

@article{wilcox-etal-2023-testing,
    title = "Testing the Predictions of Surprisal Theory in 11 Languages",
    author = "Wilcox, Ethan G.  and
      Pimentel, Tiago  and
      Meister, Clara  and
      Cotterell, Ryan  and
      Levy, Roger P.",
    journal = "Transactions of the Association for Computational Linguistics",
    volume = "11",
    year = "2023",
    address = "Cambridge, MA",
    publisher = "MIT Press",
    url = "https://aclanthology.org/2023.tacl-1.82/",
    doi = "10.1162/tacl_a_00612",
    pages = "1451--1470",
}

@inproceedings{tsipidi-etal-2024-surprise,
    title = "Surprise! {U}niform {I}nformation {D}ensity Isn{'}t the Whole Story: Predicting Surprisal Contours in Long-form Discourse",
    author = "Tsipidi, Eleftheria  and
      Nowak, Franz  and
      Cotterell, Ryan  and
      Wilcox, Ethan  and
      Giulianelli, Mario  and
      Warstadt, Alex",
    editor = "Al-Onaizan, Yaser  and
      Bansal, Mohit  and
      Chen, Yun-Nung",
    booktitle = "Proceedings of the 2024 Conference on Empirical Methods in Natural Language Processing",
    month = nov,
    year = "2024",
    address = "Miami, Florida, USA",
    publisher = "Association for Computational Linguistics",
    url = "https://aclanthology.org/2024.emnlp-main.1047/",
    doi = "10.18653/v1/2024.emnlp-main.1047",
    pages = "18820--18836"
}

@inproceedings{tsipidi-etal-2025-harmonic,
    title = "The Harmonic Structure of Information Contours",
    author = "Tsipidi, Eleftheria  and
      Kiegeland, Samuel  and
      Nowak, Franz  and
      Xu, Tianyang  and
      Wilcox, Ethan  and
      Warstadt, Alex  and
      Cotterell, Ryan  and
      Giulianelli, Mario",
    editor = "Che, Wanxiang  and
      Nabende, Joyce  and
      Shutova, Ekaterina  and
      Pilehvar, Mohammad Taher",
    booktitle = "Proceedings of the 63rd Annual Meeting of the Association for Computational Linguistics (Volume 1: Long Papers)",
    month = jul,
    year = "2025",
    address = "Vienna, Austria",
    publisher = "Association for Computational Linguistics",
    url = "https://aclanthology.org/2025.acl-long.1527/",
    doi = "10.18653/v1/2025.acl-long.1527",
    pages = "31636--31659",
    ISBN = "979-8-89176-251-0"
}

@inproceedings{degen2013cost,
  title={Cost-based pragmatic inference about referential expressions},
  author={Degen, Judith and Franke, Michael and Jager, Gerhard},
  booktitle={Proceedings of the Annual Meeting of the Cognitive Science Society},
  volume={35},
  number={35},
  year={2013}
}

@article{bergen2016pragmatic,
  title={Pragmatic reasoning through semantic inference},
  author={Bergen, Leon and Levy, Roger and Goodman, Noah},
  journal={Semantics and Pragmatics},
  volume={9},
  pages={20--1},
  year={2016}
}

@inproceedings{cohn-gordon-etal-2019-incremental,
    title = "An Incremental Iterated Response Model of Pragmatics",
    author = "Cohn-Gordon, Reuben  and
      Goodman, Noah  and
      Potts, Christopher",
    editor = "Jarosz, Gaja  and
      Nelson, Max  and
      O{'}Connor, Brendan  and
      Pater, Joe",
    booktitle = "Proceedings of the Society for Computation in Linguistics ({SC}i{L}) 2019",
    year = "2019",
    url = "https://aclanthology.org/W19-0109/",
    doi = "10.7275/cprc-8x17",
    pages = "81--90"
}

@inproceedings{white2020learning,
  title={Learning to refer informatively by amortizing pragmatic reasoning},
  author={White, Julia and Mu, Jesse and Goodman, Noah D},
  booktitle={Proceedings of the Annual Meeting of the Cognitive Science Society},
  volume={42},
  year={2020}
}

@article{oh2023why,
    author = {Oh, Byung-Doh and Schuler, William},
    title = "Why Does Surprisal From Larger Transformer-Based Language Models Provide a Poorer Fit to Human Reading Times?",
    journal = {Transactions of the Association for Computational Linguistics},
    volume = {11},
    year = {2023},
    month = {03},
    issn = {2307-387X},
    doi = {10.1162/tacl_a_00548},
    url = {https://doi.org/10.1162/tacl\_a\_00548},
    eprint = {https://direct.mit.edu/tacl/article-pdf/doi/10.1162/tacl\_a\_00548/2075940/tacl\_a\_00548.pdf},
}

@article{
shain2024evidence,
author = {Cory Shain  and Clara Meister  and Tiago Pimentel  and Ryan Cotterell  and Roger Levy },
title = {Large-scale evidence for logarithmic effects of word predictability on reading time},
journal = {Proceedings of the National Academy of Sciences},
volume = {121},
number = {10},
pages = {e2307876121},
year = {2024},
doi = {10.1073/pnas.2307876121},
URL = {https://www.pnas.org/doi/abs/10.1073/pnas.2307876121},
eprint = {https://www.pnas.org/doi/pdf/10.1073/pnas.2307876121}
}

@inproceedings{kuribayashi-etal-2024-psychometric,
    title = "Psychometric Predictive Power of Large Language Models",
    author = "Kuribayashi, Tatsuki  and
      Oseki, Yohei  and
      Baldwin, Timothy",
    editor = "Duh, Kevin  and
      Gomez, Helena  and
      Bethard, Steven",
    booktitle = "Findings of the Association for Computational Linguistics: NAACL 2024",
    month = jun,
    year = "2024",
    address = "Mexico City, Mexico",
    publisher = "Association for Computational Linguistics",
    url = "https://aclanthology.org/2024.findings-naacl.129/",
    doi = "10.18653/v1/2024.findings-naacl.129",
    pages = "1983--2005"
}

@article{farquhar2024detecting,
  title={Detecting hallucinations in large language models using semantic entropy},
  author={Farquhar, Sebastian and Kossen, Jannik and Kuhn, Lorenz and Gal, Yarin},
  journal={Nature},
  volume={630},
  number={8017},
  pages={625--630},
  year={2024},
  publisher={Nature Publishing Group UK London}
}

@inproceedings{franke2014typical,
  title={Typical use of quantifiers: A probabilistic speaker model},
  author={Franke, Michael},
  booktitle={Proceedings of the Annual Meeting of the Cognitive Science Society},
  volume={36},
  number={36},
  year={2014}
}

@inproceedings{meister-etal-2024-towards,
    title = "Towards a Similarity-adjusted Surprisal Theory",
    author = "Meister, Clara  and
      Giulianelli, Mario  and
      Pimentel, Tiago",
    editor = "Al-Onaizan, Yaser  and
      Bansal, Mohit  and
      Chen, Yun-Nung",
    booktitle = "Proceedings of the 2024 Conference on Empirical Methods in Natural Language Processing",
    month = nov,
    year = "2024",
    address = "Miami, Florida, USA",
    publisher = "Association for Computational Linguistics",
    url = "https://aclanthology.org/2024.emnlp-main.921/",
    doi = "10.18653/v1/2024.emnlp-main.921",
    pages = "16485--16498"
}

@inproceedings{giulianelli-etal-2024-generalized,
    title = "Generalized Measures of Anticipation and Responsivity in Online Language Processing",
    author = "Giulianelli, Mario  and
      Opedal, Andreas  and
      Cotterell, Ryan",
    editor = "Al-Onaizan, Yaser  and
      Bansal, Mohit  and
      Chen, Yun-Nung",
    booktitle = "Findings of the Association for Computational Linguistics: EMNLP 2024",
    month = nov,
    year = "2024",
    address = "Miami, Florida, USA",
    publisher = "Association for Computational Linguistics",
    url = "https://aclanthology.org/2024.findings-emnlp.682/",
    doi = "10.18653/v1/2024.findings-emnlp.682",
    pages = "11648--11669"
}

@article{giulianelli2026incremental,
  title = {Incremental alternative sampling as a lens into the temporal and representational resolution of linguistic prediction},
  journal = {Journal of Memory and Language},
  volume = {148},
  OPTpages = {104715},
  year = {2026},
  url = {https://www.sciencedirect.com/science/article/pii/S0749596X25001081},
  author = {Mario Giulianelli and Sarenne Wallbridge and Ryan Cotterell and Raquel Fern{\'a}ndez}
}

@inproceedings{giulianelli-etal-2023-comes,
    title = "What Comes Next? {E}valuating Uncertainty in Neural Text Generators Against Human Production Variability",
    author = "Giulianelli, Mario  and
      Baan, Joris  and
      Aziz, Wilker  and
      Fern{\'a}ndez, Raquel  and
      Plank, Barbara",
    editor = "Bouamor, Houda  and
      Pino, Juan  and
      Bali, Kalika",
    booktitle = "Proceedings of the 2023 Conference on Empirical Methods in Natural Language Processing",
    month = dec,
    year = "2023",
    address = "Singapore",
    publisher = "Association for Computational Linguistics",
    url = "https://aclanthology.org/2023.emnlp-main.887/",
    doi = "10.18653/v1/2023.emnlp-main.887",
    pages = "14349--14371"
}

@misc{hu2023expectations,
      title={Expectations over Unspoken Alternatives Predict Pragmatic Inferences}, 
      author={Jennifer Hu and Roger Levy and Judith Degen and Sebastian Schuster},
      year={2023},
      eprint={2304.04758},
      archivePrefix={arXiv},
      primaryClass={cs.CL},
      url={https://arxiv.org/abs/2304.04758}, 
}

@inproceedings{gay-etal-2026-information,
    title = "Is Information Density Uniform when Utterances are Grounded on Perception and Discourse?",
    author = "Gay, Matteo  and
      Haley, Coleman  and
      Giulianelli, Mario  and
      Ponti, Edoardo",
    editor = "Demberg, Vera  and
      Inui, Kentaro  and
      Marquez, Llu{\'i}s",
    booktitle = "Proceedings of the 19th Conference of the {E}uropean Chapter of the {A}ssociation for {C}omputational {L}inguistics (Volume 1: Long Papers)",
    month = mar,
    year = "2026",
    address = "Rabat, Morocco",
    publisher = "Association for Computational Linguistics",
    url = "https://aclanthology.org/2026.eacl-long.178/",
    doi = "10.18653/v1/2026.eacl-long.178",
    pages = "3825--3853",
    ISBN = "979-8-89176-380-7"
}

\appendix

\section{Data Preparation}
\label{app:methods-data-preprocessing}

The target utterances and their full preceding conversations are extracted from the SWDA corpus \citep{stolcke-etal-2000-dialogue}. The target utterances are then split into a context and a continuation, and the dialogue history is stored. A turn is defined as the unit of dialogue consisting of all adjacent utterances spoken by the same speaker. This approach cuts out backchannels and disfluencies in the corpus. Table~\ref{tab:processingdatastats} displays relevant dataset statistics before and after the pre-processing steps were taken.\looseness-1

\begin{table}[h]
    \centering
    \resizebox{\linewidth}{!}{%
    \begin{tabular}{lcc}
        \textbf{Dataset} & \textbf{Utterances} & \textbf{Dialogues} \\
        \hline
        Before Pre-Processing & 221,616 & 1,155 \\
        After Pre-Processing & 1,342 & 680 \\
    \end{tabular}
    }
    \caption{Data statistics of the corpus before and after pre-processing steps.}
    \label{tab:processingdatastats}
\end{table}

\subsection{Dialogue Data Cleaning and Preparation}
\label{app:methods-data-cleaning}
All of the text in the corpora was filtered through the use of regular expressions to remove any formatting or tags in the utterances. For example: <<motorcycle noise>>. Additionally, regular expressions were also used to remove repeated words, interrupted words, and short disfluencies such as ``um'' or ``uh''. Additional annotations of contextual noise in the conversations were also removed. Short utterances that are backchannels are then removed, these are utterances that contain three or less words. The dataset contains several cases where adjacent utterances within a dialogue were spoken by the same speaker. This was usually because of a long pause or interruption in the dialogue. The previous step of removing backchannels also leads to more adjacent utterances by the same speaker. Adjacent utterances are therefore combined in order to keep these continuous singular thoughts together. The corpus was then re-indexed and numbered so that there were no missing or duplicated values in the identification columns. 

\subsection{Extracting Target Utterances}
\label{app:methods-data-extract-targets}
Not all of the utterances in the dataset were considered high quality enough to be used for this experiment. In order to select target utterances, the following criteria were devised: the context must be of length 3 or greater to allow for enough of the dialogue history that the model can attempt to predict the intent of the speaker, the utterance must have a verb as this is the location where the utterances are split, and the utterance must be between 10 and 30 words to guarantee that utterances will be split into contexts that are long enough to provide the language model with enough information to generate continuations without confusing the models with unnecessary or extra information.
Target utterances are selected as the first sentence in a turn, which ensures that they are most directly related to the information provided by the previous turn.

\subsection{Extracting Dialogue History}
\label{app:methods-data-extract-contexts}
We are interested in the preceding dialogue history before the context of each utterance. For each target utterance we extract the preceding turn spoken by the other speaker, and the maximum number of utterances from the preceding dialogue history that can fully fit within the 1024 token limit of the GPT-2, the model used for surprisal calculation.

\subsection{Choice Points: Splitting Sentences}
\label{app:methods-data-splitting-targets}
Inspired by Jaeger's classic \textit{that}-mentioning example, we opt for the matrix verb as our choice point. Matrix verbs mark a point at which the speaker is likely to have already developed an utterance plan and they must make a choice as to how to realise their intended completion. The fact that some meaningful partial sentence context has been established constrains the alternatives to a degree (rather than simply predicting the continuation from ``the boy''), but not to an extent that the continuation is necessarily obvious (e.g., ``the cat sat on the'').

We thus split utterances at the root verb to create the utterance context and the human continuation. Root verbs are identified using spaCy's part-of-speech tagger and dependency parser.\footnote{\href{https://spacy.io/}{https://spacy.io/}}

Other choice points would be feasible and interesting to study under the same framework, and we are excited about future work that examines this more exhaustively. Choice points could be chosen, for example, by tracking other syntactic structures. For instance, dative verbs (like ``the man gave \ldots'') could be completed with either a prepositional object structure (``\ldots the book to the boy'') or with a double object dative structure (``\ldots the boy the book''). Alternatively, other more linear criteria could be used—for example, choice points could be determined by measuring continuation entropy at every position in the sentence and choosing the positions with the highest entropy.

\section{Generating Alternatives}
\label{app:generating-alternatives}

Language models are used as the simulators of dialogue production in order to generate alternative continuations. The model used to generate alternatives was OpenAI's GPT-4o\footnote{\href{https://platform.openai.com/docs/models/gpt-4o}{https://platform.openai.com/docs/models/gpt-4o}}.
In order to ensure that the model generated continuations in the correct format, a one-shot prompting method was used to provide the model with a single simple demonstration of a sentence completion. 

\subsection{Goal-Agnostic Alternatives}
\label{app:generating-alternatives-completion}

The sentence continuation generation methods prompt the language model to generate continuations of the target context. This was done by providing the model with the context and various levels of context. The models' objective was defined through the use of the developer prompt, specifying to complete the provided context. 

Each method of sentence continuation was provided with the same developer prompt.

{

\ex. [Developer:] Your task is to complete the provided sentence. Complete the sentence in a natural manner, as if engaging in a phone call conversation. Only write the continuation to the sentence without any additional information or words in your response.

\ex. [User:] Complete the sentence: ``The cat jumped"

\ex. [Assistant:] The cat jumped over the dog.

}

\paragraph{No Dialogue History.}
The no history condition provides the model with the context but no other information. This gives the model the most freedom to complete the sentence.\\

\noindent No Dialogue History User Prompt

\ex. [User:] Complete the sentence: ``\{context\}"

\paragraph{Previous Utterance.}
The preceding history condition provides the model with the context and the directly preceding utterance.
\\

\noindent Previous Utterance History User Prompt

\ex. [User:] Given this sentence from speaker A: ``\{history\}", Complete the sentence from Speaker B: ``\{context\}"

\paragraph{All Previous Utterances.}
The full history conditions provide the model with all previous utterances in the dialogue, up to a maximum of 1000 tokens.
\\

\noindent Full History User Prompt

\ex. [User:] Given this phone conversation between Speaker A and Speaker B: ``\{history\}", Complete the sentence from Speaker \{SpeakerID\}: ``\{context\}"

\subsection{Goal-Directed Alternatives}
\label{app:generating-alternatives-paraphrasing}
To generate goal-directed alternatives we prompt GPT-4o to paraphrase the full sentence, but only allow it to change the words after the context. This approach leverages the language model's internal representation of the semantic closeness of words. In order to limit the repetition of paraphrases, all of the paraphrases for a target utterance are generated at once and specified to be unique from each other. This method resulted in several occurrences where the model refused to generate the paraphrases at all, or refused to generate all 10 paraphrases. For example the model would return ``I'm sorry but I don't understand." or ``I can't do that." In order to keep the data balanced, it was decided to drop all continuations of the contexts that had failed to produce the 10 paraphrases. This reduced the number of contexts to 309 with 309 human continuations, and 12,360 continuations representing 3,090 continuations of each generation method, or 10 continuations per method per context. This is the final dataset upon which the calculations were performed.\\

\noindent Paraphrase Generation Developer Prompt
\ex. [Developer]: Your task is to paraphrase the provided sentence. Paraphrase the sentence in a natural manner, as if engaging in a phone call conversation, while still keeping the sentence factually the same. Only write the paraphrase of the sentence without any additional information or words in your response. Try your best to do so even if the provided text seems nonsensical or does not have proper sentence structure.\\

\vspace{-1em}
\noindent Paraphrase Generation User Prompt
\ex. [User:] Write 10 unique paraphrases of the following sentence: ``\{text\}", the paraphrases must always start with the following words: ``\{context\}"\\

\vspace{-1em}
\noindent Paraphrase Generation Assistant Prompt

\ex. [User:] Write 10 unique paraphrases of the following sentence: ``The cat jumped over the dog.", the paraphrases must always start with the following words: ``The cat jumped over"
    
\ex. [Assistant:] The cat jumped over the husky.\\
    The cat jumped over the dog without making a sound.\\
    The cat jumped over the dog as it wagged its tail.\\
    The cat jumped over the dog, barely missing it.\\
    The cat jumped over the dog and landed smoothly.\\
    The cat jumped over the dog while it was dozing off.\\
    The cat jumped over the dog to escape quickly.\\
    The cat jumped over the dog in a playful manner.\\
    The cat jumped over the dog and continued running.\\
    The cat jumped over the dog, surprising it.

\subsubsection{Paraphrase Detection}
\label{app:detecting-paraphrases}
In order to determine which generations should be kept and which should be removed, an analysis of the contexts combined with the generated continuations compared to the original human sentences was performed.
Following \citet{farquhar2024detecting}, we initially made use of an MNLI model to detect paraphrases and semantic equivalence, but through manual inspection, found this approach to be less effective than GPT-4, with a 85\% vs. 98\% success rate on a manually-annotated sample of 400 items.
We tested two Natural Language Inference models trained on the MNLI corpus~\citep{williams2018broadcoveragechallengecorpussentence},
Roberta Large MNLI~\citep{liu2019robertarobustlyoptimizedbert} and FLAN-T5 Base MNLI~\citep{chung2022scalinginstructionfinetunedlanguagemodels}, chosen for their high scores on the MNLI benchmark.\footnote{\href{https://paperswithcode.com/sota/natural-language-inference-on-multinli}{https://paperswithcode.com/sota/natural-language-inference-on-multinli}} The labels produced were either entailment, neutral, or contradiction, we take entailment as goal conditioned, and merge neutral and contradiction for goal agnostic. However, upon manual inspection and annotation of a sample of 400 generations, we abandoned this approach, since there was too high a level of ambiguity in terms of the neutral label, and too high an error rate. We instead opt for prompting GPT-4o to detect whether two sentences are paraphrases, a well-defined and more simple task (especially for the relatively short sentences that we select as targets), which exploits LMs greater level of capability for semantic meaning over logical equivalency \citep{mirzadeh2025gsmsymbolicunderstandinglimitationsmathematical}, as well as using a far larger and more powerful model. The following prompts were used for the paraphrase detection with GPT-4o.
\\

\vspace{-1em}
\noindent Paraphrase Detection Developer Prompt
\ex. [Developer:] Your task is to determine whether or not two sentences are paraphrases of each other. You are to classify the sentences into one of two labels: ``yes" if the sentences are paraphrases or ``no" if they are not. Do not provide any explanation for your choice, just the name of the label.\\
\vspace{-1em}

Paraphrase Detection User Prompt
\ex. [User:] Classify whether these sentences are paraphrases. Sentence A: ``\{text\}", Sentence B: ``\{generation\}"
\\

\vspace{-1em}
Paraphrase Detection Assistant Prompt
\ex. [User:] Classify whether these sentences are paraphrases. Sentence A: ``The cat jumped over the dog.", Sentence B: ``The cat jumped over a dog."

\ex. [Assistant:] Yes

Through the use of this model a set of goal-directed alternatives was created. This is the set of alternatives that are considered paraphrases by the language model.

\section{Statistical Analyses}
\subsection{Poisson–binomial Test for Rank-1 Outcomes}
\label{app:poisson–binomial}
We provide here the full specification of the test used to assess whether human utterances are ranked first more often than expected by chance.

For each trial $i \in \{1,\dots,N\}$, let $n_i$ denote the number of alternatives in the relevant alternative set. Let $Y_i \in \{0,1\}$ be a Bernoulli random variable indicating whether the human utterance is ranked first under the cost measure of interest. 
Under the null hypothesis $H_0$ of random selection among alternatives, the probability of a rank-1 outcome on trial $i$ is
\begin{equation}
P(Y_i = 1 \mid H_0) = \frac{1}{n_i}.
\end{equation}

The total number of rank-1 outcomes across trials is given by the random variable
\begin{equation}
K = \sum_{i=1}^{N} Y_i,
\end{equation}
which follows a Poisson--binomial distribution corresponding to the sum of independent Bernoulli variables with heterogeneous success probabilities $\{1/n_i\}_{i=1}^{N}$.
Further let
\begin{equation}
K_{\mathrm{obs}} = \sum_{i=1}^{N} y_i
\end{equation}
denote the observed number of rank-1 outcomes in the data, where $y_i$ is the realised value of $Y_i$. Statistical significance is assessed using a one-sided test,\looseness-1
\begin{equation}
p = P(K \ge K_{\mathrm{obs}} \mid H_0),
\end{equation}
which quantifies the probability of observing at least as many rank-1 outcomes as in the data under the null hypothesis of chance selection.

We employ a one-sided test because cost minimisation predicts an excess, but not a deficit, of rank-1 outcomes relative to chance.

\subsection{Pairwise Logistic Choice Model}
\label{app:pairwise-logit}

This appendix provides a detailed description of the pairwise logistic choice model used in \Cref{sec:condlogit}.

\paragraph{Data construction.}
For each context, we construct binary comparisons between the human continuation and each candidate alternative. Let $\alternative_i$ denote the human continuation and $\alternative_j$ an alternative from the same context. Each pair yields a binary outcome $y_{ij} \in {0,1}$, where $y_{ij}=1$ indicates that the human continuation is preferred over the alternative. 
To obtain a balanced dataset, we include both $(i,j)$ and $(j,i)$.

\paragraph{Predictors.}
For each cost function $\cost \in \{ \costsurp, \costuidlocal, \costuidglobal, \costlen \}$, as introduced in \cref{sec:costs}, we define a cost difference variable
\begin{align}
\Delta \cost_{ij} = \cost(\alternative_i; \ctx) - \cost(\alternative_j; \ctx),
\end{align}
which captures the relative cost of the human continuation compared to the alternative. 
Cost differences are standardised to have mean zero and unit variance prior to estimation. We also include a binary indicator $\mathrm{GD}_j \in \{0,1\}$, where $\mathrm{GD}_j = 1$ if the alternative $\alternative_j$ is drawn from the goal-directed set and $\mathrm{GD}_j = 0$ if it is drawn from the goal-agnostic set.

\paragraph{Model specification.}
We model the probability that the human continuation is preferred using a logistic regression:
\begin{align}
\log \frac{P(y_{ij}=1)}{1 - P(y_{ij}=1)}
=&\ \beta_0 + \beta_1 \,\Delta \cost_{ij}
+ \beta_2 \,\mathrm{GD}_{j} \nonumber \\ 
&+ \beta_3 \,(\Delta \cost_{ij} \times \mathrm{GD}_{j})
\end{align}
The coefficient $\beta_1$ captures cost sensitivity in the goal-agnostic condition, while $\beta_3$ captures the change in cost sensitivity in the goal-directed condition. The total cost sensitivity under goal-directed alternatives is therefore given by $\beta_1 + \beta_3$.
As mentioned in \Cref{sec:condlogit}, this specification corresponds to the two-alternative restriction of the softmax production rule. 
In this interpretation, the slope on $\Delta \cost_{ij}$ estimates the negative cost-sensitivity parameter $- \alpha$, such that more negative coefficients indicate stronger sensitivity to a given cost. 

\paragraph{Estimation.}
Models are estimated via maximum likelihood. Standard errors are clustered at the context level to account for the dependence induced by multiple pairwise comparisons within the same context. For each cost function, we fit a separate model and report coefficient estimates, confidence intervals, and per-observation log-likelihoods.

\paragraph{Interpretation.}
A negative coefficient on $\Delta \cost_{ij}$ indicates that lower-cost continuations are more likely to be selected. 
The interaction term allows us to test whether this effect differs between goal-directed and goal-agnostic alternatives within a single unified model.

\section{Additional Results}
\label{app:results}

\subsection{Lexical Overlap between Context and Continuation}
\label{sec:lexica-overlap}

\begin{figure}[h!]
    \centering
    \includegraphics[width=\linewidth]{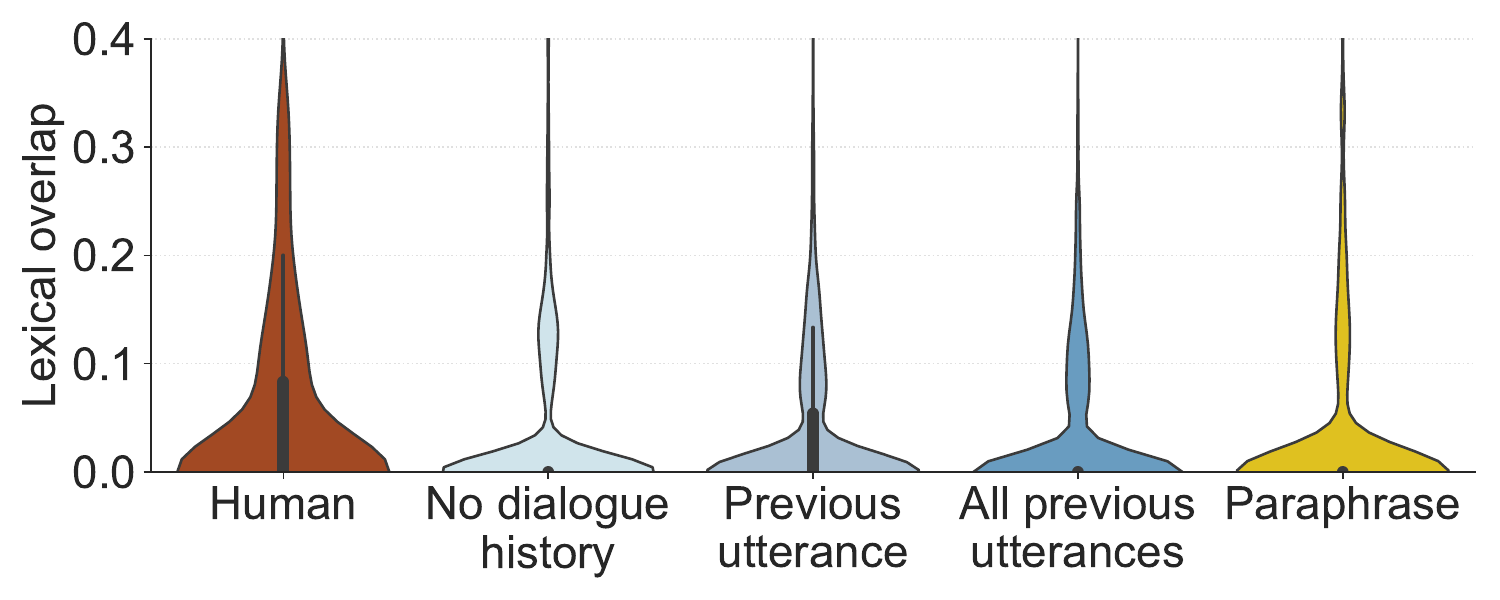}
    \caption{Distribution of lexical overlap between context and continuations by continuation type.}
    \label{fig:lex_overlap}
\end{figure}

We evaluate lexical overlap between context and continuations across human and LM-generated alternatives, similar to~\citet{molnar2023attribution}. The results of these comparisons can be found in \cref{fig:lex_overlap}.

\subsection{Goal Predictability from Context}
\label{sec:goal-predictability-from-context}
Goal-agnostic alternative sets are intended to capture a listener's uncertainty over both the speaker's intended goal and its realisation. In practice, however, contextual constraints often make some goals more predictable than others. As a result, even without conditioning explicitly on the goal, a listener may assign non-zero probability to continuations that align with the speaker's intended goal.

\begin{figure}[h!]
    \centering
    \includegraphics[width=\linewidth]{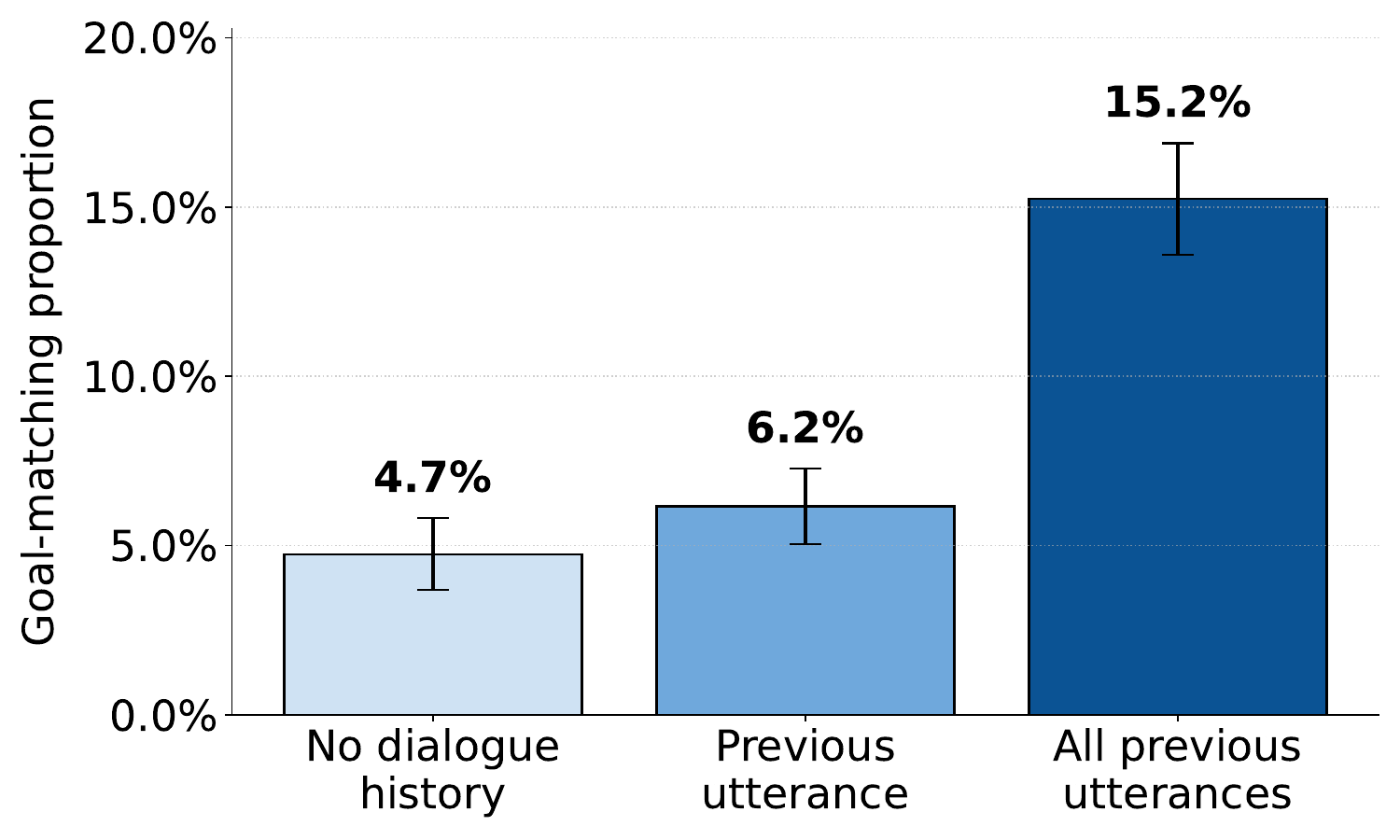}
    \caption{Proportion of goal-agnostic alternatives that match the goal of the observed utterance, across dialogue history conditions.}
    \label{fig:goal_match_proportions}
\end{figure}

We examine this in our data by identifying goal-agnostic continuations that are also contained in the goal-directed set. We find a small but non-zero overlap, which increases with additional context: with longer context windows, a larger proportion of goal-agnostic continuations match the intended goal. \Cref{fig:goal_match_proportions} breaks down these proportions.

\begin{figure*}[ht!]
    \centering
    \includegraphics[width=\linewidth]{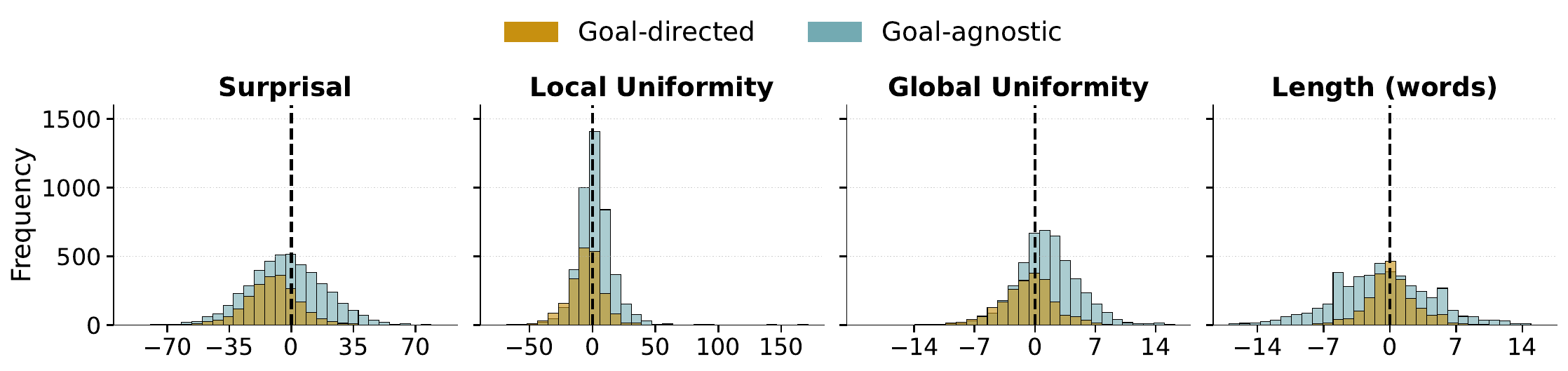}
    \caption{Pairwise cost differences between human continuations and alternatives, computed as the cost of the human continuation minus the cost of the alternative. Negative values indicate that the human continuation has lower cost than the alternative (i.e., is preferred under the cost), while positive values indicate that the alternative has lower cost.}
    \label{fig:delta_diffs_dists}
    \vspace{-0.5em}
\end{figure*}

\subsection{Cost Distribution Differences by Context}
\label{app:cost_diff_dists}
We additionally examine per-context differences in cost between human continuations and sampled goal-directed and goal-agnostic alternatives. Under the model in \Cref{sec:theory}, speakers are expected to prefer continuations with lower cost than competing alternatives. To operationalise this, we sample a single alternative per context and compute its cost difference relative to the human continuation.
The resulting difference distributions are shown in \Cref{fig:delta_diffs_dists}.

One-sided t-tests (\Cref{tab:ttest_results_diffs}) show that surprisal exhibits a strong effect under both the goal-directed and the goal-agnostic conditions, with a markedly larger effect in the goal-directed case. Local uniformity and global uniformity show a significant effect only in the goal-directed comparison, with differences in the opposite direction, and thus non-significant under the one-sided test, in the goal-agnostic condition. The opposite pattern holds for length.
Overall, these results confirm that surprisal most clearly distinguishes human continuations from goal-directed alternatives.

\begin{table}[h!]
\centering
\small
\resizebox{\linewidth}{!}{%
\begin{tabular}{lcc|cc}
\toprule
 & \multicolumn{2}{c|}{\textbf{Goal-directed}} & \multicolumn{2}{c}{\textbf{Goal-agnostic}} \\
\textbf{Cost difference} & $t$ & $p$-value & $t$ & $p$-value \\
\midrule
Surprisal & $-32.48$ & $< 10^{-8}$ & $-8.23$ & $< 10^{-8}$ \\
Local uniformity & $-13.37$ & $< 10^{-8}$ & \textcolor{gray}{$10.57$} & \textcolor{gray}{$1.00$} \\
Global uniformity & $-12.11$ & $< 10^{-8}$ & \textcolor{gray}{$26.48$} & \textcolor{gray}{$1.00$} \\
Length (words) & \textcolor{gray}{$2.99$} & \textcolor{gray}{$1.00$} & $-10.72$ & $< 10^{-8}$ \\
\bottomrule
\end{tabular}
}
\caption{Results of one-sided t-tests comparing observed productions to alternative continuations, testing whether mean differences are smaller than zero.}
\label{tab:ttest_results_diffs}
\vspace{-0.5em}
\end{table}

\subsection{Pairwise Logistic Choice Model}
\Cref{tab:pairwise_logit_results} contains supporting details of coefficients and confidence intervals for \Cref{fig:logit_coefficients_loglik}. 
The pairwise logistic choice model and the interpretation of the coefficients is presented in \Cref{app:pairwise-logit}.

\begin{table*}[h]
\centering
\resizebox{\linewidth}{!}{%
\begin{tabular}{l cc cc cc c}
\toprule
& \multicolumn{2}{c}{\textbf{Goal-agnostic}} 
& \multicolumn{2}{c}{\textbf{Goal-directed}} 
& \multicolumn{2}{c}{\textbf{Interaction}} \\
\cmidrule(lr){2-3} \cmidrule(lr){4-5} \cmidrule(lr){6-7}
\textbf{Cost} 
& $\beta_1$ & CI 
& $\beta_1 + \beta_3$ & CI 
& $\beta_3$ & CI 
& \textbf{Per-item LL} \\
\midrule

Surprisal 
& $-0.304^{\dagger}$ & $[-0.498, -0.110]$
& $-2.073$ & $[-2.525, -1.622]$
& $-1.769$ & $[-2.210, -1.328]$
& $-0.615$ \\

Local uniformity 
& $~~~0.357$ & $[~~~0.156, ~~~0.559]$
& $-0.683^{\dagger}$ & $[-1.195, -0.170]$
& $-1.040$ & $[-1.532, -0.548]$
& $-0.670$ \\

Global uniformity 
& $~~~0.796$ & $[~~~0.593, ~~~0.999]$
& $-0.632$ & $[-0.859, -0.405]$
& $-1.428$ & $[-1.678, -1.179]$
& $-0.637$ \\

Length 
& $-0.325$ & $[-0.503, -0.148]$
& \textcolor{gray}{$~~~0.233$} & \textcolor{gray}{$[-0.113, ~~~0.579]$}
& $~~~0.558$ & $[~~~0.250, ~~~0.867]$
& $-0.681$ \\

\bottomrule
\end{tabular}%
}
\caption{Pairwise logistic regression estimates per cost measure. Results are significant at $p<0.001$, daggers ($\dagger$) indicate coefficients with $p < 0.01$, \textcolor{gray}{gray} indicates non-significance.
\looseness-1
}
\label{tab:pairwise_logit_results}
\end{table*}

\subsection{Conditional Logit Model of Graded Cost Sensitivity}
\label{app:conditional_logit}

\begin{table*}[h!]
\centering
\resizebox{\linewidth}{!}{%
\begin{tabular}{lcccc@{\hspace{1em}}cccc}
\toprule
& \multicolumn{4}{c}{\textbf{Goal-directed alternatives}}
& \multicolumn{4}{c}{\textbf{Goal-agnostic alternatives}} \\
\cmidrule(lr){2-5}\cmidrule(lr){6-9}
\textbf{Cost}
& $\alpha$ & $\ell (\uparrow)$ & $P(\text{rank}\!=\!1)$ & $P(\text{best vs 2nd})$
& $\alpha$ & $\ell (\uparrow)$ & $P(\text{rank}\!=\!1)$ & $P(\text{best vs 2nd})$ \\
\midrule
Surprisal
& \textbf{1.918} & \textbf{-0.187} & \textbf{0.464} & \textbf{0.701}
& 0.253$^\dagger$ & -0.173 & 0.090 & 0.518 \\
Local uniformity
& 0.825 & -0.235 & 0.245 & 0.557
& -0.359 & -0.172 & 0.111 & 0.549 \\
Global uniformity
& 0.748 & -0.241 & 0.223 & 0.552
& \textbf{-0.840} & \textbf{-0.164} & \textbf{0.169} & \textbf{0.584} \\
Length (words)
& -0.335$^\ddagger$ & -0.250 & 0.171 & 0.526
& 0.339 & -0.172 & 0.098 & 0.521 \\
\midrule
Uniform \textit{(baseline)}
& -- & -2.018 & 0.142 & 0.500
& -- & -2.727 & 0.067 & 0.500 \\
\bottomrule
\end{tabular}}
\caption{Conditional logit results for production choice in goal-directed and goal-agnostic alternative sets, reporting the estimated cost-sensitivity coefficient $\alpha$, the average per-item log-likelihood $\ell$, the expected probability $P(\text{rank}\!=\!1)$ that the human continuation is ranked first under the corresponding cost measure, and the probability $P(\text{best vs 2nd})$ that the cost-minimising alternative is preferred over the second-best alternative. Daggers ($\dagger$) indicate coefficients with $p < 0.01$; double daggers ($\ddagger$) indicate coefficients with $p < 0.05$; all other coefficients are significant at $p < 0.001$.}
\label{tab:logit_model_results}
\end{table*}

In addition to our analyses in the main body of the paper, we also model utterance choice as probabilistic using a conditional logit model.
For each context $\ctx_i$ in the dataset, let $\altset{i}$ denote the corresponding alternative set---either the goal-agnostic set $\altset{\ctx_i}$ or the goal-directed set $\altset{\ctx_i, \goal}$ as defined in \Cref{sec:theory-alternative-sets}. Let $Y_i$ be a categorical random variable over $\altset{i}$ representing the production choice in context $\ctx_i$. 
Each alternative $\alternative \in \altset{\ctx_i}$ is associated with a cost $\cost \in \{ \costsurp, \costuidlocal, \costuidglobal, \costlen \}$, as introduced in \cref{sec:costs}.
We model the probability of selecting continuation $\alternative$ from $\altset{i}$ as
\begin{align}
\label{eq:condlogit}
P(Y_i = \alternative \mid\!\altset{\ctx_i})
\;=\;
\frac{\exp\!\left(- \alpha \, \cost(\alternative; \ctx_i)\right)}
{\sum_{\alternative' \in \altset{\ctx_i}}
\exp\!\left(- \alpha \, \cost(\alternative'; \ctx_i)\right)}, 
\end{align}
where $\alpha$ is a scalar coefficient corresponding to the sensitivity parameter of the choice model in \Cref{sec:theory-choice-model}.

We fit conditional logit models with a single cost measure at a time, standardising each cost across the dataset, and compare them using four metrics: (1) the estimated cost-sensitivity coefficient~$\alpha$; (2) the average per-item log-likelihood~$\ell$; (3) the expected probability that the model assigns to the lowest-cost alternative, $P(\text{rank}\!=\!1)$, which describes how confidently the model selects the top alternative (in other words, when $P(\text{rank}\!=\!1) = 1$, the model selects the top alternative  deterministically); and (4) the probability that the cost-minimising alternative is preferred over the second-best, $P(\text{best vs.\ 2nd})$. This last metric is comparable across goal-directed and goal-agnostic conditions, as it is insensitive to alternative set size.
\Cref{tab:logit_model_results} summarises these results.

Within goal-directed alternative sets, surprisal is by far the strongest predictor of choice, with a large positive cost-sensitivity coefficient, the highest log-likelihood, and a  $P(\text{rank}\!=\!1)$ substantially higher than chance.
This model also achieves the highest $P(\text{best vs.\ 2nd})$ across the board, confirming speaker-oriented surprisal minimisation as the account that best explains production choice.

Uniformity-based costs also show an asymmetry across alternative sets. In goal-directed contexts, both local and global uniformity significantly predict choice.
In contrast, within goal-agnostic sets both measures have significant negative coefficients, indicating that greater uniformity reduces the likelihood of selection when evaluated against listener-available alternatives.
Utterance length shows the weakest effect within goal-directed alternatives, where it has a negative coefficient, and remains comparatively weak relative to uniformity costs in goal-agnostic sets. 

\subsection{Results without Stratified Sampling}
\label{sec:app-no-stratification}
\begin{figure*}[h!]
    \centering
    \includegraphics[width=\linewidth]{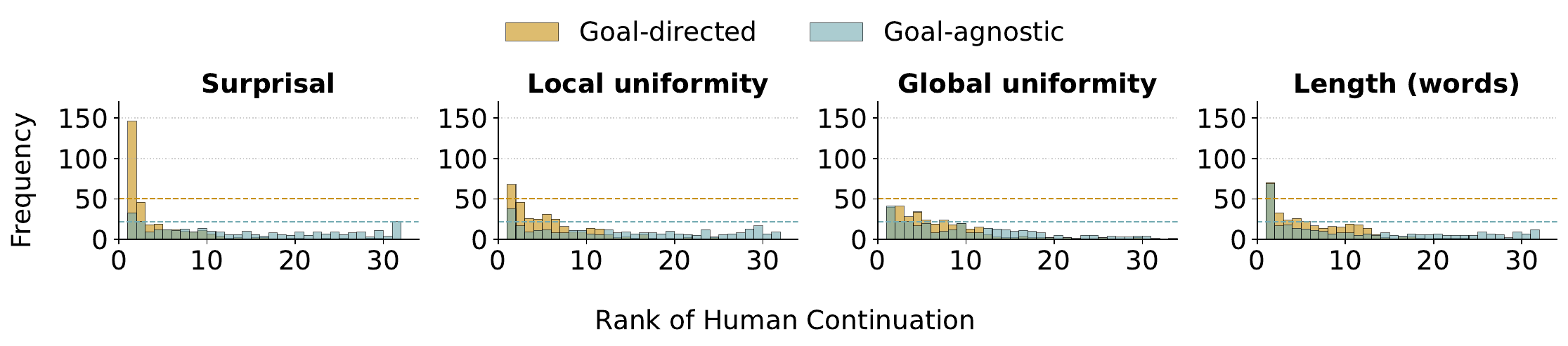}
    \caption{
    \textit{Results without Stratified Sampling}. Ranking distributions of observed human continuations under different cost measures, evaluated against goal-directed and goal-agnostic alternative sets. A rank of 1 indicates that the human continuation has the lowest cost among the available alternatives, corresponding to deterministic cost minimisation.
    }
    \label{fig:rankingbymetric_nostratification}
\end{figure*}

\begin{figure*}[t]
    \centering
    \includegraphics[width=0.95\textwidth]{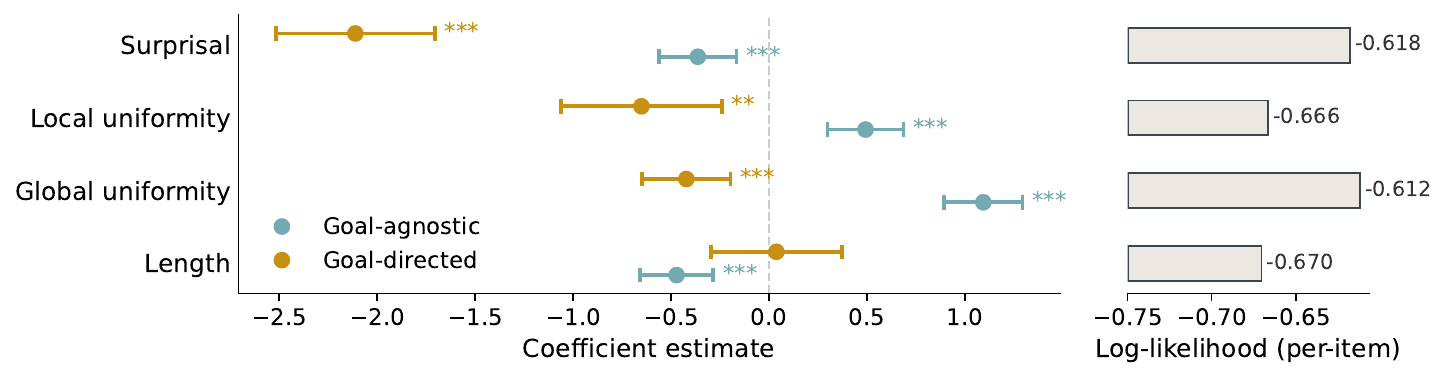}
    \caption{
    \textit{Results without Stratified Sampling}. Logistic regression results predicting whether a continuation is selected over an alternative as a function of the difference in cost with respect to the alternative, goal condition, and their interaction. Points show coefficient estimates on the log-odds scale, horizontal bars indicate 95\% confidence intervals. Asterisks indicate significance levels ($p < .05$, $p < .01$, $p < .001$). On the right panel is the corresponding per-item log-likelihood for each model.
    }
    \label{fig:logit_coefficients_loglik_nostratification}
\end{figure*}

In this section, we report the results obtained using the full set of generated utterances, without applying the stratified sampling procedure described in \Cref{sec:stratified-sampling}. The results, shown in \Cref{fig:rankingbymetric_nostratification,tab:poisson_binomial_results_nostratification,fig:logit_coefficients_loglik_nostratification,tab:pairwise_logit_results_nostratification}, are qualitatively aligned to those reported in the main text. One numerical difference is in the log-likelihood of the global uniformity model, which is larger in the results without stratified sampling. This is likely due to the global distributional differences that we remove through stratification.

\begin{table}[h]
\vspace{2.3em}
\centering
\resizebox{\linewidth}{!}{%
\begin{tabular}{l c|c}
\toprule
\textbf{Cost} & \textbf{Goal-directed} & \textbf{Goal-agnostic} \\
\midrule
Surprisal 
& 47.6\% {\scriptsize\textcolor{YlOrBr8}{$\bf\times 5.12$}} 
& 10.7\% {\scriptsize\textcolor{YlOrBr5}{$\bf\times 3.24$}} \\
Local uniformity 
& 22.1\% {\scriptsize\textcolor{YlOrBr4}{$\bf\times 2.38$}} 
& 12.4\% {\scriptsize\textcolor{YlOrBr6}{$\bf\times 3.76$}} \\
Global uniformity 
& \textcolor{gray}{13.0\% {\scriptsize$\bf\times 1.40$}} 
& 13.4\% {\scriptsize\textcolor{YlOrBr7}{$\bf\times 4.06$}} \\
Length (words) 
& 22.8\% {\scriptsize\textcolor{YlOrBr4}{$\bf\times 2.45$}} 
& 22.5\% {\scriptsize\textcolor{YlOrBr9}{$\bf\times 6.82$}} \\
\midrule
Uniform \textit{(baseline)}  & 9.3\% & 3.3\% \\
\bottomrule
\end{tabular}%
}
\caption{\textit{Results without Stratified Sampling}. Percentage of observed rank-1 outcomes, with multiplicative increase relative to the uniform baseline. All values exceed the uniform baseline under one-sided Poisson--binomial tests ($p < 10^{-8}$ or smaller), except for global uniformity against goal-directed alternatives ($p = 0.018$), which does not show a robust effect.}
\label{tab:poisson_binomial_results_nostratification}
\end{table}

\begin{table*}[h]
\centering
\resizebox{\linewidth}{!}{%
\begin{tabular}{l cc cc cc c}
\toprule
& \multicolumn{2}{c}{\textbf{Goal-agnostic}} 
& \multicolumn{2}{c}{\textbf{Goal-directed}} 
& \multicolumn{2}{c}{\textbf{Interaction}} \\
\cmidrule(lr){2-3} \cmidrule(lr){4-5} \cmidrule(lr){6-7}
\textbf{Cost} 
& $\beta_1$ & CI 
& $\beta_1 + \beta_3$ & CI 
& $\beta_3$ & CI 
& \textbf{Per-item LL} \\
\midrule

Surprisal 
& $-0.364$ & $[-0.561, -0.167]$
& $-2.111$ & $[-2.517, -1.705]$
& $-1.747$ & $[-2.147, -1.347]$
& $-0.618$ \\

Local uniformity 
& $~~~0.492$ & $[~~~0.298, ~~~0.685]$
& $-0.651^{\dagger}$ & $[-1.062, -0.241]$
& $-1.143$ & $[-1.531, -0.756]$
& $-0.666$ \\

Global uniformity 
& $~~~1.093$ & $[~~~0.893, ~~~1.292]$
& $-0.423$ & $[-0.648, -0.198]$
& $-1.515$ & $[-1.753, -1.278]$
& $-0.612$ \\

Length 
& $-0.472$ & $[-0.660, -0.285]$
& \textcolor{gray}{$~~~0.037$} & \textcolor{gray}{$[-0.298, ~~~0.372]$}
& $~~~0.509$ & $[~~~0.219, ~~~0.799]$
& $-0.670$ \\

\bottomrule
\end{tabular}%
}
\caption{\textit{Results without Stratified Sampling}. Pairwise logistic regression estimates per cost measure. Results are significant at $p<0.001$, daggers ($\dagger$) indicate coefficients with $p < 0.01$, \textcolor{gray}{gray} indicates non-significance.
\looseness-1
}
\label{tab:pairwise_logit_results_nostratification}
\end{table*}

\end{document}